\documentclass[10pt,twocolumn,letterpaper]{article}

\usepackage[pagenumbers]{cvpr}

\usepackage[dvipsnames]{xcolor}

\usepackage{graphicx}
\usepackage{makecell}
\usepackage{enumitem}
\usepackage{color, colortbl}
\usepackage{tikz}
\usepackage{subcaption}
\usepackage{wrapfig}
\usepackage{multirow}
\usepackage{listings}
\usepackage{algorithm}
\usepackage{algorithmic}
\usepackage{stfloats}

\definecolor{tabhighlight}{HTML}{e5e5e5}

\newcommand{\method}{{\color[RGB]{0,0,0}\text{SCEdit}}\xspace}

\definecolor{cvprblue}{rgb}{0.21,0.49,0.74}
\usepackage[pagebackref,breaklinks,colorlinks,citecolor=cvprblue]{hyperref}

\begin{document}

\crefname{algorithm}{Alg.}{Algs.}
\Crefname{algorithm}{Algorithm}{Algorithms}

\title{SCEdit: Efficient and Controllable Image Diffusion \\ Generation via Skip Connection Editing}

\author{
\textbf{Zeyinzi Jiang} \quad \textbf{Chaojie Mao} \quad \textbf{Yulin Pan} \quad \textbf{Zhen Han} \quad \textbf{Jingfeng Zhang} \\
Alibaba Group\\
{\tt\small \{zeyinzi.jzyz, chaojie.mcj, yanwen.pyl, hanzhen.hz, zhangjingfeng.zjf\}@alibaba-inc.com}
\vspace{-10pt}
}

\twocolumn[{
\maketitle
\begin{center}
\centering
\vspace{-11pt}
\includegraphics[width=1.0\linewidth]{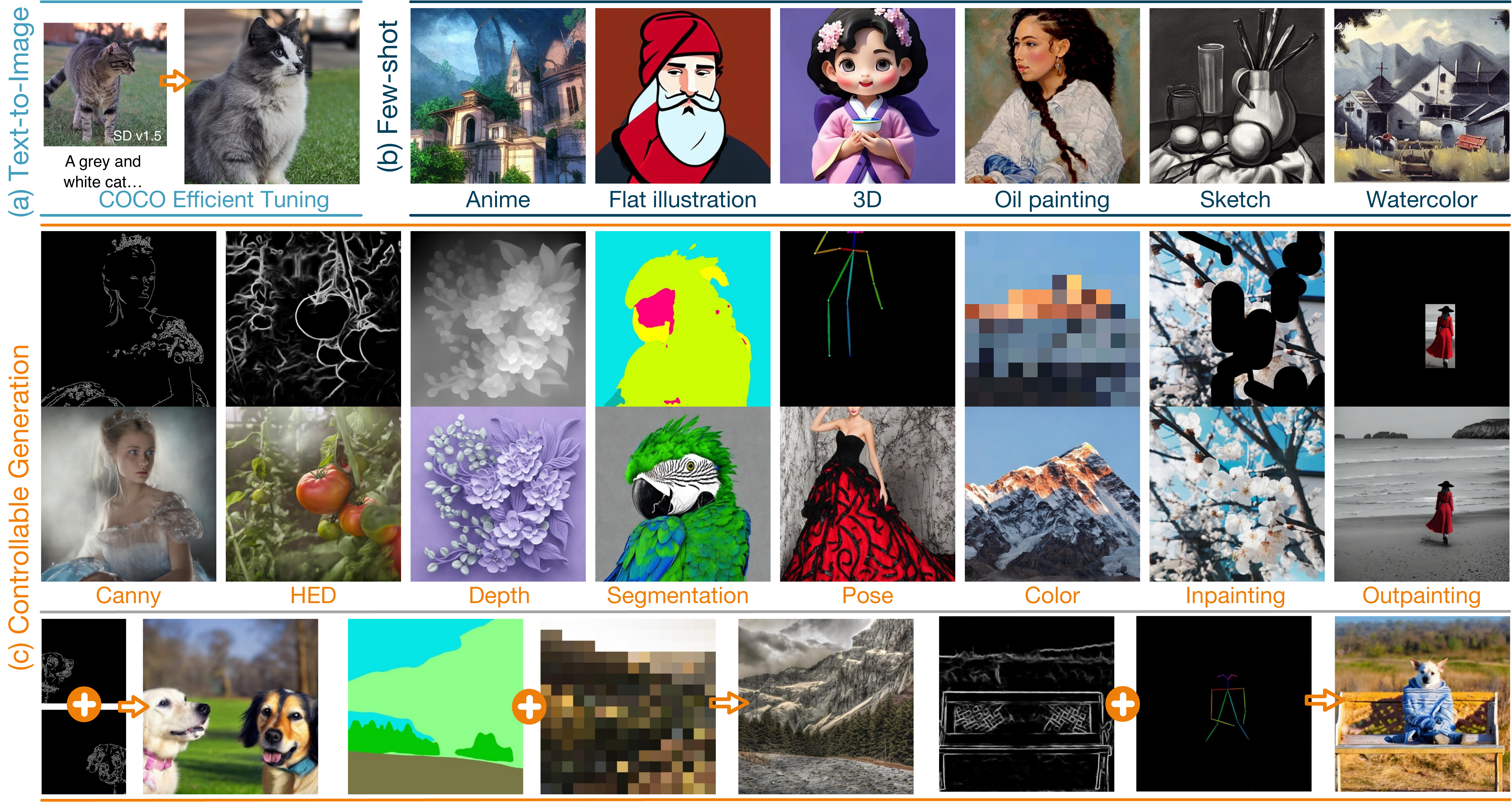}
\vspace{-14pt}
\captionsetup{type=figure}
\caption{
    \textbf{Images generated by SCEdit.} 
    With a small number of trainable parameters and low memory usage, SCEdit enables efficient fine-tuning on specific datasets (left top) and supports transfer learning with a few samples (right top).    
    Additionally, it adopts various conditions as inputs for efficient controllable generation (middle), while individually learned conditional models combine effortlessly, providing endless compositional possibilities (bottom).
}
\label{figa:show}
\vspace{-3pt}
\end{center}
}]

\begin{abstract}

Image diffusion models have been utilized in various tasks, such as text-to-image generation and controllable image synthesis.
Recent research has introduced tuning methods that make subtle adjustments to the original models, yielding promising results in specific adaptations of foundational generative diffusion models. 
Rather than modifying the main backbone of the diffusion model, we delve into the role of skip connection in U-Net and reveal that hierarchical features aggregating long-distance information across encoder and decoder make a significant impact on the content and quality of image generation. 
Based on the observation, we propose an efficient generative tuning framework, dubbed \textbf{\texttt{SCEdit}}, which integrates and edits Skip Connection using a lightweight tuning module named SC-Tuner. 
Furthermore, the proposed framework allows for straightforward extension to controllable image synthesis by injecting different conditions with Controllable SC-Tuner, simplifying and unifying the network design for multi-condition inputs.
Our SCEdit substantially reduces training parameters, memory usage, and computational expense due to its lightweight tuners, with backward propagation only passing to the decoder blocks.
Extensive experiments conducted on text-to-image generation and controllable image synthesis tasks demonstrate the superiority of our method in terms of efficiency and performance.
Project page: \url{https://scedit.github.io/}

\end{abstract}
\section{Introduction}
\label{sec:intro}

Building upon large-scale pre-trained image diffusion models~\cite{dalle2,sd15,sd21,glide}, researchers have focused on various downstream tasks and applications, including text-to-image generation~\cite{wanx,midjourney,dalle3,pixart,muse}, controllable image synthesis ~\cite{composer,controlnet,t2i-adapter} and image editing~\cite{dreambooth,cyclenet,sdedit,ip2p}. 
However, fully fine-tuning a foundation image diffusion model often proves inefficient or even impractical in most customized scenarios due to the constraints of limited training data and computational resources.

\begin{figure}[t]
    \centering
    \includegraphics[width=1.0\linewidth]{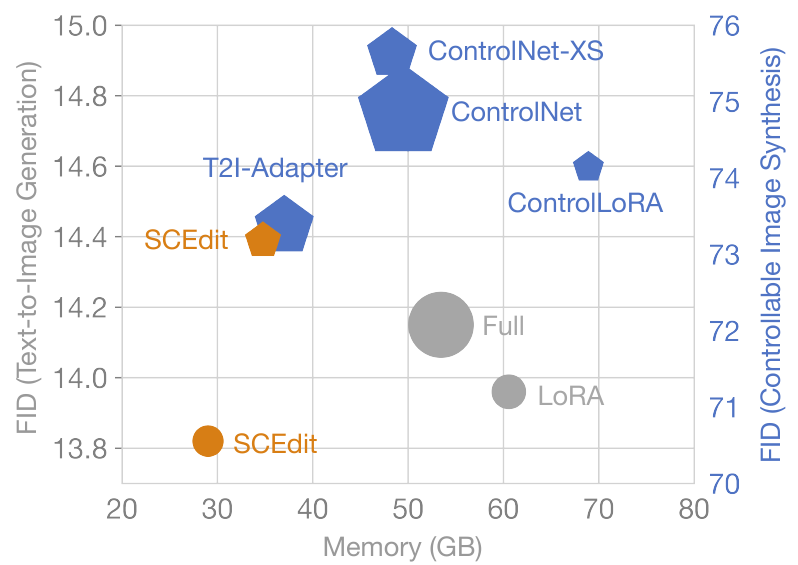}
    \vspace{-15pt}
    \caption{\textbf{Performance and efficiency comparison} on both text-to-image generation (circular markers) and controllable image synthesis (pentagonal markers) tasks. The marked area reflects the relative amount of parameters.
    }
    \label{figa:efficient}
    \vspace{-5pt}
\end{figure}

Recently, efficient tuning methods~\cite{difffit,restuning,utuning} have emerged as a practical solution by introducing additional trainable structures on generative tasks. 
Nonetheless, many of these popular efficient tuning methods still suffer from substantial resource consumption as the network expands. 
For instance, LoRA~\cite{lora} typically adds the trainable low-rank matrices to multi-head attention layers all across the U-Net~\cite{unet}, and backward propagation is conducted throughout the entire backbone, resulting in an accumulation of gradients and increase in memory usage during training, even more than that in fully fine-tuning. 
To address this issue, we properly design our framework by strategically inserting all the trainable modules into the Skip Connections (SCs) between the encoder and decoder of the U-Net, effectively decoupling the encoder from the backpropagation process and significantly reducing computation and memory requirements, as shown in ~\cref{figa:efficient}.

\begin{figure}[ht]
    \centering
    \includegraphics[width=1.0\linewidth]{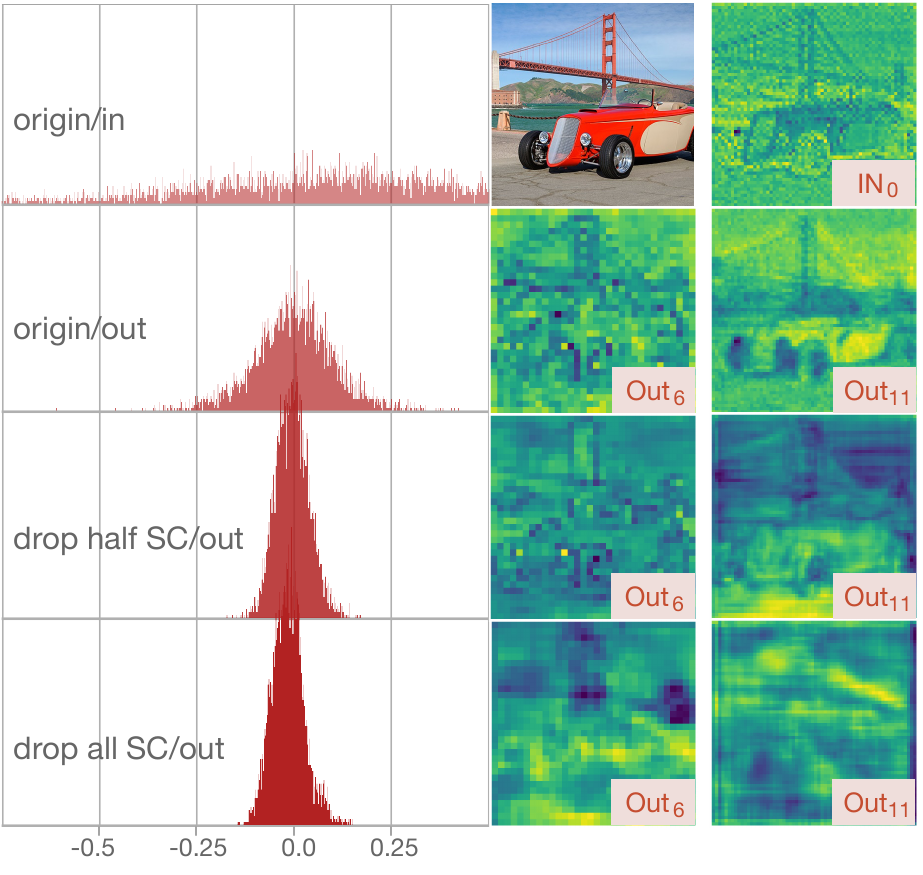}
    \vspace{-15pt}
    \caption{The \textbf{output distributions} (left) and \textbf{feature maps} (right) of pre-trained U-Net. Removing the skip connections from different layers markedly affects the overall network output.}
    \label{figa:motivation}
    \vspace{-20pt}
\end{figure}

The SCs bridge the gap between distant blocks in the U-Net architecture, facilitating the integration of information over long distances and alleviating vanishing gradients. Some works~\cite{scalelong,sde,sr3} have focused on exploring the efficacy of SCs in enhancing the training stability of the U-Net, as well as in improving the generation quality~\cite{freeu}.
Inspired, we further investigate the potential of U-Net's SCs in adapting to new scenarios.
To gain a deeper insight into each SC within a pre-trained U-Net, we gradually remove SCs and observe the subsequent changes in the value distributions and feature maps across the blocks of the decoder in ~\cref{figa:motivation}.
We first assess the diversity of information in the final decoder's output by visualizing the distribution of its latent feature values. A higher variance signifies a greater breadth of information, while a variance near zero indicates a significant loss of detail. As we discard an increasing number of SC from half to all, the variance of the final decoder output gradually decreases. 
In addition, when half of the SCs are removed, there is a marked reduction in the level of detailed structural information within the feature maps of both the 6th and 11th decoder blocks. Eliminating all SCs further intensifies the deterioration of information.
These trends further validate the significant impact of SC on the generation of detailed structural information.

%%%%%%%%%%%%%%%%%%%%%%%%%%%%%%%%%%%%%%%%
%%%%%%%%%%%%%%%%%%%%%%%%%%%%%%%%%%%%%%%%

%%%%%%%%%%%%%%%%%%%%%%%%%%%%%%%%%%%%%%%%
%%%%%%%%%%%%%%%%%%%%%%%%%%%%%%%%%%%%%%%%

In light of this revelation, we propose a simple yet highly efficient approach by \textbf{S}kip \textbf{C}onnection \textbf{Edit}ing for image generation, dubbed \textbf{\texttt{\method}}. 
Specifically, we introduce a lightweight tuning module named \textbf{\texttt{SC-Tuner}}, which is designed to edit the latent features within each SC of the pre-trained U-Net for efficient tuning. 
Furthermore, we extend the capabilities of SC-Tuner to facilitate controllable image synthesis, which can accommodate various input conditions using the Controllable SC-Tuner (\textbf{\texttt{CSC-Tuner}}).
Our SCEdit framework is adaptable to a broad spectrum of image generation tasks using the proposed tuning modules and by decoupling the encoder blocks in U-Net, it allows for efficient and flexible training, as it enables backward propagation solely through the decoder block.

We evaluate our \method on efficient tuning of text-to-image generation tasks as well as controllable image synthesis tasks. 
For text-to-image generation tasks, our approach outperforms existing methods on COCO2017 dataset in terms of FID score and qualitative results, while also reducing memory consumption by 52\% during the training stage. 
Additionally, faster transfer and high-quality results are achieved in the few-shot fine-tuning scenario.
On the controllable generation tasks, our approach can easily control the results under various conditional inputs and show impressive results, while having lower computational costs than existing methods. It utilizes merely 7.9\% of the parameters required by ControlNet and achieves a 30\% reduction in memory usage.
\section{Related work} 
\label{sec:related}

\noindent\textbf{Image Diffusion Models}~\cite{ddpm,sde,ddim} have achieved the state-of-the-art performance in sample quality of generative image. 
Research on text-to-image diffusion models~\cite{ldm,dalle2} incorporates text latent vectors as conditions. Some works ~\cite{controlnet,composer} conduct controllable image generation by leveraging various images as conditions, offering personalization, customization, or task-specific image generation.
However, the scale and computational cost limit the applications of image diffusion models.

\noindent\textbf{Efficient Tuning}~\cite{lora, adapter, prompt, hypernetworks, restuning} has become a popular solution in recent years, attracting significant attention. 
It allows for easy adaptation of image diffusion models by making light adjustments to the original pre-trained models.
LoRA~\cite{lora} is to use low-rank matrices to learn weight offsets, which has proven effective in customized image generation scenarios. 
Other works, such as U-Tuning~\cite{utuning}, Res-Tuning~\cite{restuning}, and MAM-Adapter~\cite{mamadapter}, propose a unified paradigm for efficient tuning methods, offering more options for tuning pre-trained models. 
Typically, some tuning modules are integrated into the U-Net architecture of diffusion models, but few works thoroughly analyze the key factors for improving the quality of generative images through tuning modules.

\noindent\textbf{U-Net} is originally introduced for diffusion models in DDPM~\cite{ddpm}, and it achieves tremendous success in the field of generative tasks.
Recently, there has been a lot of attention focused on the effectiveness of skip connections in U-Net. 
To address the issue of doubling signal variance caused by skip connections and alleviate the oscillations of U-Net, some methods~\cite{sde,sr3} propose rescaling all skip connections by $\frac{1}{\sqrt{2}}$. 
ScaleLong~\cite{scalelong} provides further theoretical analysis on why scaling the coefficients of skip connections can help improve the training stability of U-Net and using constant scaling or learnable scaling for better stability.
FreeU~\cite{freeu} recognizes that the main backbone of U-Net plays a primary role in denoising, while the skip connections introduce high-frequency features into the decoder module. By leveraging the strengths of both components, the denoising capability of U-Net can be enhanced. 
Inspired by previous works, we aim to analyze the roles of skip connections in the image generation process and propose a novel and effective framework incorporating tuning modules by editing skip connections.
\section{Method}
\label{sec:method}

\subsection{Preliminaries}
\label{sec:prel}

%%%%%%%%%%%%%%%%%%%%%%%%%%%%%%%%%%%%%%%%
\begin{figure*}[t]
    \includegraphics[width=1.0\linewidth]{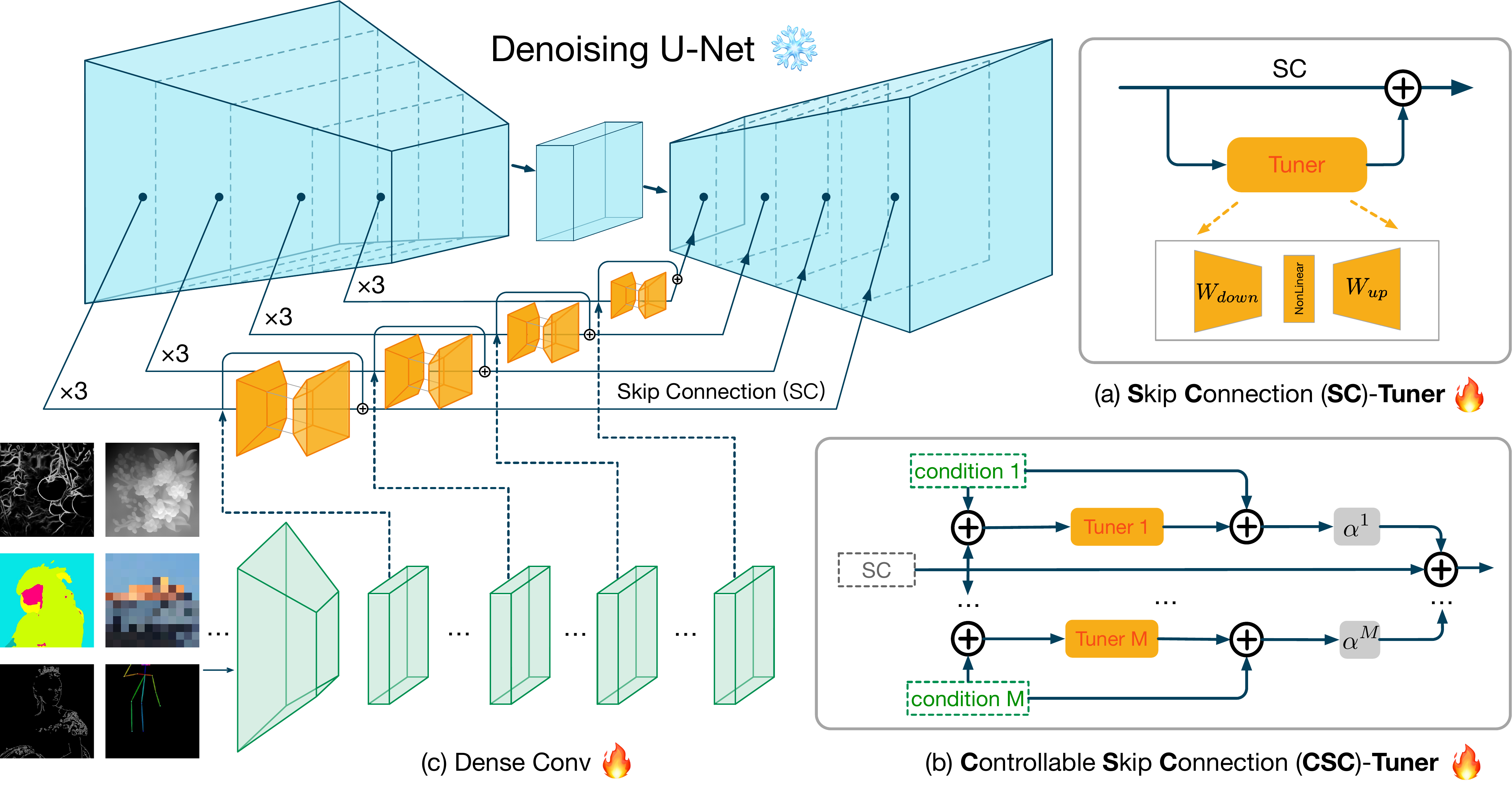}
    \caption{\textbf{Illustration of SCEdit framework.} 
    Our method achieves efficient tuning by editing the features on skip connections which is proved to have rich structural information. 
    Leveraging (a) SC-Tuner for text-to-image generation tuning, and controllable image synthesis can be achieved with the assistance of (b) CSC-Tuner and (c) Cascade Dense Convolution.
    }
    \vspace{-12pt}
    \label{figa:framework}
\end{figure*}
%%%%%%%%%%%%%%%%%%%%%%%%%%%%%%%%%%%%%%%%

\noindent 
Diffusion models~\cite{ddpm} are a family of probabilistic generative models that aim at sampling high-fidelity images from Gaussian noise. It combines two fundamental processes: a diffusion process and a denoising process. In the diffusion process, it gradually decreases the signal-to-noise ratio of an image via a $T$-step Markov chain, following prescribed noise level schedules $[\beta_{1}, \beta_{2}, ..., \beta_{T}]$. At each step $t$, the noisy intermediate variable $x_{t}$ is constructed as:
%%%%%%%%%%%%%%%%%%%%%%%%%%%%%%%%%%%%%%%%
\begin{equation}
    x_{t} = \sqrt{\Bar{\alpha}_{t}}x_{0} + \sqrt{1-\Bar{\alpha}_{t}}\epsilon, \quad \epsilon \sim \mathcal{N}(0, 1),
\end{equation}
%%%%%%%%%%%%%%%%%%%%%%%%%%%%%%%%%%%%%%%%
\noindent where $\alpha_{t} = 1 - \beta_{t}$ and $\Bar{\alpha}_{t} = \prod_{s=1}^{t}\alpha_{s}$. The denoising process reverses the above diffusion process by estimating the noise $\epsilon$ via a parameterized neural network, which is trained by minimizing the $l_2$ loss between estimated noise $\epsilon_{\theta}$ and real noise $\epsilon$:
%%%%%%%%%%%%%%%%%%%%%%%%%%%%%%%%%%%%%%%%
\begin{equation}
    \ell_{\text{simple}}^{t}(\theta) = \mathbb{E}_{x_{0},t,\epsilon}\left\Vert \epsilon_{\theta}(x_{t}, t) - \epsilon \right\Vert_{2}^{2}.
\end{equation}
%%%%%%%%%%%%%%%%%%%%%%%%%%%%%%%%%%%%%%%%
\noindent 
U-Net~\cite{unet} is already a widely adopted architecture in pixel-corresponded generative tasks, such as image restoration, medical segmentation, and the aforementioned image generation. Specifically, it first encodes the input into multi-scale features through multiple cascaded encoder blocks: 
%%%%%%%%%%%%%%%%%%%%%%%%%%%%%%%%%%%%%%%%
\begin{equation}
    x_{i+1} = \mathcal{F}_{i}(x_{i}), \quad 0 \leq i \leq N-1,
\end{equation}
%%%%%%%%%%%%%%%%%%%%%%%%%%%%%%%%%%%%%%%%
where $i$ denotes the index number of the encoder layer, $\mathcal{F}_{i}$ denotes the $i$-th encoder block operation, $N$ is the total number of encoder layers, $x_{i+1}$ denotes the output of $i$-th encoder layer and $x_{0}$ means the original image input. 

Subsequently, the decoder gradually decodes the feature via establishing a skip connection with corresponding encoder layer outputs, which complements the high-frequency information:
%%%%%%%%%%%%%%%%%%%%%%%%%%%%%%%%%%%%%%%%
\begin{equation}
    g_{j+1} = \mathcal{G}_{j}([x_{N-j}; g_{j}]), \quad 0 \leq j \leq N-1,
\end{equation}
%%%%%%%%%%%%%%%%%%%%%%%%%%%%%%%%%%%%%%%%
where $[\cdot;\cdot]$ represents the concatenation operation, $j$ denotes the index number of the decoder layer, $\mathcal{G}_{j}$ denotes the $j$-th decoder block operation, $N$ is the total number of decoder layers which equals that of encoder layers, $g_{j+1}$ denotes the output of $j$-th decoder layer, $g_{0}$ is the last output block before decoder input.

\subsection{Tuner modules}
\label{sec:tuner}

We present \textbf{S}kip \textbf{C}onnection \textbf{Tuner}, termed \textbf{\texttt{SC-Tuner}}, a method designed to directly edit the latent features within skip connections.
As illustrated in ~\cref{figa:framework}{\color{red}a}, the SC-Tuner is composed of a tuning operation, referred to as Tuner OP, and a residual connection.
The $j$-th SC-Tuner takes $x_{N-j}$ as input and produces the sum of $x_{N-j}$ and the output of Tuner OP with $x_{N-j}$ as input. 
This process can be mathematically formulated as follows:
%%%%%%%%%%%%%%%%%%%%%%%%%%%%%%%%%%%%%%%%
\begin{equation}
\label{eqa:sctuner}
    \text{O}^{SC}_{j}(x_{N-j}) = \mathcal{T}_{j}(x_{N-j})+x_{N-j}, \\
\end{equation}
%%%%%%%%%%%%%%%%%%%%%%%%%%%%%%%%%%%%%%%%
where $\text{O}^{SC}_{j}(x_{N-j})$ denotes the $j$-th SC-Tuner module with $x_{N-j}$ as input, $\mathcal{T}_{j}$ denotes the Tuner OP of $j$-th SC-Tuner module.

The efficient tuning paradigm~\cite{restuning} has demonstrated comparable effectiveness across various Tuner OPs, including LoRA OP, Adapter OP, and Prefix OP, as well as their versatile combinations.
In these independent OPs, we adapt the form of an Adapter OP, as it has been proven to be the simplest yet relatively effective method.
The Tuner OP $\mathcal{T}_{j}$ can be defined as follows:
%%%%%%%%%%%%%%%%%%%%%%%%%%%%%%%%%%%%%%%%
\begin{equation}
    \mathcal{T}_{j}(x_{N-j}) = \mathbf{W}^{up}_{j}\phi(\mathbf{W}^{down}_{j}x_{N-j}),
\end{equation}
%%%%%%%%%%%%%%%%%%%%%%%%%%%%%%%%%%%%%%%%
where $\mathbf{W}^{up}$ and $\mathbf{W}^{down}$ are up and down tunable projection matrices, respectively, $\phi$ is a GELU~\cite{gelu} activation function. Formally, we can employ tuners of various types and scales to modify the features of skip connections. 

Furthermore, The SC-Tuner can be easily adapted to support controllable image synthesis by incorporating the conditions and $x_{N-j}$ as input, where conditions accommodate both single and multiple conditions.
In the case of multi-condition controllable generation, we assign weights to different condition branches and combine them with original skip connections. 
This modified structure is depicted in ~\cref{figa:framework}{\color{red}b}, named \textbf{C}ontrollable \textbf{SC-Tuner} or \textbf{\texttt{CSC-Tuner}}. 
We extend ~\cref{eqa:sctuner} by injecting extra conditional information which can be formulated as follows:
%%%%%%%%%%%%%%%%%%%%%%%%%%%%%%%%%%%%%%%%
% \begin{equation}
\vspace{-5.5pt}
\begin{align}
    \text{O}^{CSC}_{j}(x_{N-j}, C_{j}) = 
    \sum\limits_{m=1}^{M} & \alpha^{m}(\mathcal{T}^{m}_{j}(x_{N-j} + 
    c^{m}_{j}) + c^{m}_{j}) \nonumber \\
    & + x_{N-j},
\end{align}
% \vspace{3pt}
% \end{equation}
%%%%%%%%%%%%%%%%%%%%%%%%%%%%%%%%%%%%%%%%
where $\text{O}^{CSC}_{j}(x_{N-j}, C_{j})$ denotes the $j$-th CSC-Tuner module with $x_{N-j}$ and $M$ conditions $C_{j}=\{c^{0}_{j}, \dots, c^{M}_{j}\}$ as inputs, $c^{m}_{j}$ is the $j$-th hint block's output of ${m}$-th condition features, $\alpha^{m}$ is the weight of different independent condition embeddings, and $\sum \nolimits_{m=1}^{M}\alpha^{m}=1$. 
In practice, we can engage in joint training with multi-conditions or perform inference under combined conditions directly after completing single-condition training. 
The control conditions include but are not limited to the canny edge, depth, hed boundary, semantic segmentation, pose keypoint, color, and masked image.

\subsection{SCEdit framework}
\label{sec:scedit}

We introduce \textbf{\texttt{\method}}, a framework designed for efficient \textbf{S}kip \textbf{C}onnection \textbf{Edit}ing in image generation that utilizes the SC-Tuner and CSC-Tuner. 
All the skip connections and conditions are fed into the SC-Tuner and CSC-Tuner, with the outputs subsequently concatenated to the original feature maps and input into the corresponding decoder blocks.
We depict this process as follows:
%%%%%%%%%%%%%%%%%%%%%%%%%%%%%%%%%%%%%%%%
\begin{equation}
    g_{j+1} = \mathcal{G}_{j}([\text{O}_{j}(x_{N-j}, C_{j}); g_{j}]), \\
\end{equation}
%%%%%%%%%%%%%%%%%%%%%%%%%%%%%%%%%%%%%%%%
where $\text{O}_{j}(x_{N-j}, C_{j})$ denotes the $j$-th SC-Tuner or CSC-Tuner modules. By editing the original $x_{N-j}$, we can adapt it to different tasks.

\method can facilitate flexibility and efficiency across both text-to-image generation and controllable image synthesis tasks.
As illustrated in ~\cref{figa:framework}, the SC-Tuner is applied to the text-to-image generation task and integrated into all skip connections within the pre-trained U-Net architecture. 
The CSC-Tuner is employed for controllable image synthesis, where the input of corresponding condition features is encoded via a cascaded convolution network, as shown in ~\cref{figa:framework}{\color{red}c}, consisting of a multi-layer hint block and several dense modules for feeding into the skip connection branch, where each level contains zero convolution layers and SiLU~\cite{gelu} activation functions, ensuring compatibility with the dimensions of skip connection.
\section{Experiments} \label{sec:exp}

\subsection{Experimental setups}

\noindent
\textbf{Evaluation.}
We evaluate the flexibility and efficiency of our SCEdit, mainly through the text-to-image generation and controllable image synthesis tasks. We analyze its quantitative metrics such as trainable parameters, memory consumption, training speed, and FID~\cite{fid} evaluation metrics to assess its performance and efficiency. Additionally, we consider qualitative evaluations of the quality and fidelity of the generated images.

\noindent
\textbf{Datasets.}
In the text-to-image generation task, we use the COCO2017~\cite{coco} dataset for training and testing, which consists of 118k images and 591k training prompts. Furthermore, we employ customized style datasets~\cite{customstyle} with a limited samples to further validate the effectiveness of our approach. 
In controllable image synthesis tasks, for each kind of condition data, we utilize a filtered version of the LAION artistic dataset~\cite{laion5b} that includes approximately 600k images that have been removed of duplicates, low-resolution images, those with a risk for harm, and those that are of low quality.

\noindent
\textbf{Baselines.}
According to task categories, the evaluation tasks can be characterized into two groups: \textbf{(i)} the text-to-image tuning includes fully fine-tuning and LoRA~\cite{lora} tuning strategies; \textbf{(ii)} the controllable image synthesis focuses on methods including additional information as input, such as ControlNet~\cite{controlnet}, T2I-Adapter~\cite{t2i-adapter}, ControlLoRA~\cite{controllora}, and ControlNet-XS~\cite{controlnetxs}. Due to the distinct task characteristics of these two categories, we apply the SC-Tuner module for the former and CSC-Tuner for the latter.

\noindent
\textbf{Implementation details.}
For all experiments, we perform efficient fine-tuning based on the Stable Diffusion (SD) pre-trained model, where SD v1.5~\cite{sd15} is used for the text-to-image task and SD v2.1~\cite{sd21} for the conditional task, and the input image sizes for training are set to 512$\times$512. We utilize the AdamW~\cite{adamw} optimizer with a fixed learning rate of 5e-5. Unless otherwise specified, models are trained for 100k steps. 

Following ControlNet, we use different input conditions, \textit{e.g.}, edge map~\cite{canny,hed}, depth map~\cite{depth}, segmentation map~\cite{uniformer}, and body key points~\cite{openpose}, in addition to the color map condition in T2I-Adapter~\cite{t2i-adapter} and the mask-generation strategy presented in LaMa~\cite{lama}.

\subsection{Text-to-image generation}

%%%%%%%%%%%%%%%%%%%%%%%%%%%%%%%%%%%%%%%%
\begin{figure}[t]
    \includegraphics[width=1.0\linewidth]{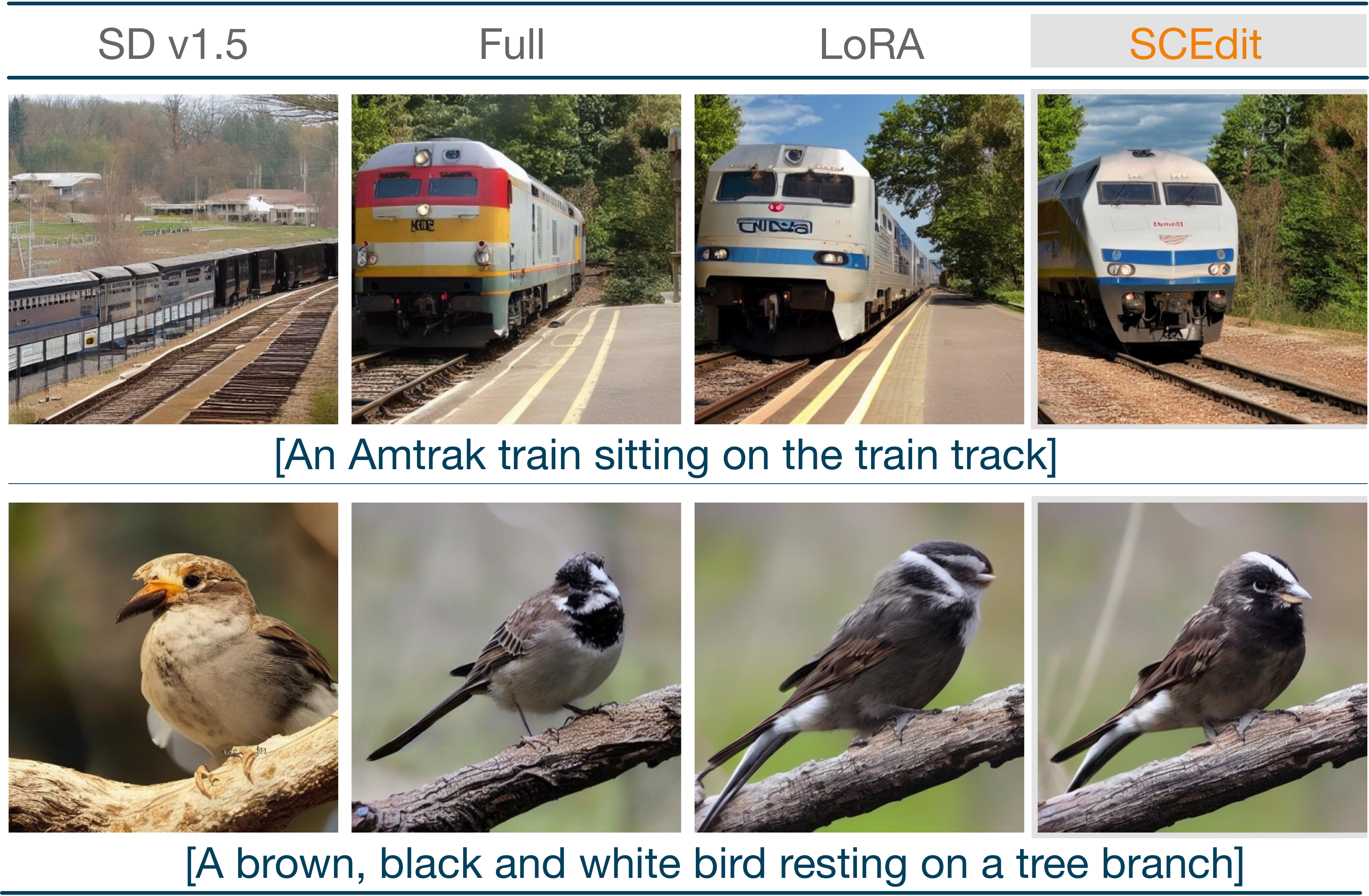}
    \caption{\textbf{Qualitative comparison} of the original SD v1.5, existing tuning strategies, and our SCEdit using the same prompts.}
    \label{figa:coco_img_cmp}
    \vspace{-5pt}
\end{figure}
%%%%%%%%%%%%%%%%%%%%%%%%%%%%%%%%%%%%%%%%

%%%%%%%%%%%%%%%%%%%%%%%%%%%%%%%%%%%%%%%%
\begin{table}[t]
\caption{
    \textbf{Comparison of FID and efficiency on text-to-image generation} using COCO2017 dataset.
    For the FID score, we follow the default settings of SD v1.5. In terms of efficiency, we compare the aspects of parameter-, memory-, and training time efficiency. We compare the performance of LoRA and our method under two different parameter settings.
}
\label{taba:coco_effi_cmp}
\vspace{2pt}
\centering
\begin{tabular}{l|c|ccc}
\toprule
Method & FID$\downarrow$ & Params & Mem. & Time \\
\midrule
SD v1.5~\cite{sd15} & 15.48 & - & - & - \\
Full & 14.15 & 859.52M & 53.46G & $\times$1.0 \\
LoRA/r=64~\cite{lora} & 13.96 & 23.94M & 60.57G & $\times$1.24 \\
LoRA/r=6~\cite{lora} & 15.12 & 2.24M & 59.94G &  $\times$1.20 \\
\midrule
SCEdit & \textbf{13.82} & 19.68M & 29.02G & $\times$0.78 \\
SCEdit$_{10}$ & 13.99 & \textbf{1.98}M & \textbf{28.06}G & \textbf{$\times$0.77} \\
\bottomrule
\end{tabular}
% \vspace{-2mm}
\end{table}

%%%%%%%%%%%%%%%%%%%%%%%%%%%%%%%%%%%%%%%%

\noindent
\textbf{Performance and efficiency comparison.}
To evaluate the transferability of our method in downstream tasks, we fine-tuning COCO2017 with pre-trained U-Net and compare our method with other training strategies in terms of qualitative and quantitative results.

The qualitative results can be seen in ~\cref{figa:coco_img_cmp}, the leftmost column is the zero-shot result of the original model, while the fine-tuned model obtained semantic comprehension capabilities on downstream tasks.
Compared to existing strategies, our method exhibits fewer artificial artifacts in the generated images and higher visual quality. For instance, in the second row, the generated bird has more realistic details in the head while maintaining semantic comprehension.

In addition, as shown in ~\cref{taba:coco_effi_cmp}, compared to the fully fine-tuning baseline, our SCEdit achieves 0.33 performance gain in FID score while using only 2.29\% of the parameters and reducing nearly 22\% training time.
It is worth noted that at lower parameters compared to \text{LoRA/r=64} strategies with rank 64, our method is lower in both FID, while the training memory can be reduced by 52.1\% and the training time of LoRA is 1.6$\times$ longer than ours.
When we further reduce the hidden dimensions of the tuner by 10$\times$, denoted as $\text{SCEdit}_{10}$, which corresponds to a reduction of parameters by the same factor compared to \text{LoRA/r=6}, our FID score shows a significant decreased by 1.13, while also maintaining a clear advantage in terms of memory usage and training time.

%%%%%%%%%%%%%%%%%%%%%%%%%%%%%%%%%%%%%%%%
\begin{figure*}[t]
    \includegraphics[width=1.0\linewidth]{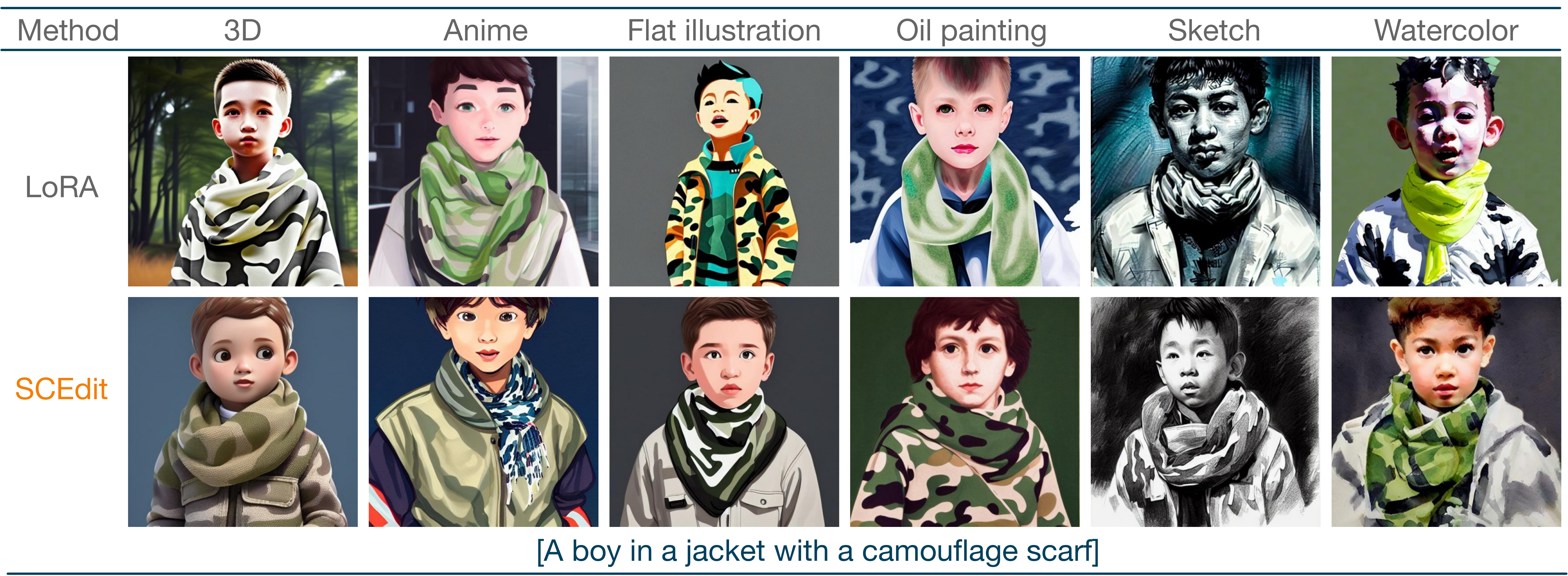}
    \caption{\textbf{Few-shot transfer learning comparison} on various custom-stylized datasets. 
    Compared to LoRA tuning, SCEdit achieves more precise learning of style characteristics and generates images of superior quality.
    }
    \label{figa:fewshot_img_cmp}
\end{figure*}
%%%%%%%%%%%%%%%%%%%%%%%%%%%%%%%%%%%%%%%%

\noindent
\textbf{Few-shot transfer learning.}
In the image generation community, few-shot learning is a practical technique that enables users to train a personalized model with just a small subset of data.
Our experiment involved performing few-shot transfer learning on a custom-stylized dataset, which included classes 3D, anime, flat illustration, oil painting, sketch, and watercolor, each with only 30 image-text pairs. 
Moreover, to prevent style leakage, we perform a cleansing of style-related words from the original prompts and incorporated specific trigger words \textless sce\textgreater\ during training to ensure the reliability of the experiment.
In ~\cref{figa:fewshot_img_cmp}, we provide a comparison of the quality of samples between LoRA and our method, employing the same training setup. 
The results demonstrate that our method more accurately captures the style, aligning with the distribution of the original training data while maintaining text alignment. For instance, the flat illustration style retained the descriptions of a camouflage scarf, while the sketch style preserved the line-drawn depictions in monochrome.

%%%%%%%%%%%%%%%%%%%%%%%%%%%%%%%%%%%%%%%%
\begin{table}[t]
\caption{
    \textbf{Comparison of FID and efficiency on controllable image synthesis} using LAION dataset.
    ``k'' denotes the convolution kernel size of conditional model, and a larger size performs better on FID score, albeit with a moderate increase in parameters.
    }
\label{taba:control_effi_cmp}
% \vspace{2pt}
\setlength{\tabcolsep}{4pt}
\centering
\begin{tabular}{l|c|ccc}
\toprule
Method & FID$\downarrow$ & Params. & Mem. & Time \\
\midrule
ControlNet~\cite{controlnet} & 74.86 & 364.23M & 49.51G & $\times$1.0 \\
T2I-Adapter~\cite{t2i-adapter} & 73.37 & 77.37M & 37.01G & $\times$0.92 \\
ControlNet-XS~\cite{controlnetxs} & 75.63 & 55.30M & 48.32G & $\times$0.97 \\
ControlLoRA~\cite{controllora} & 74.14 & \textbf{21.52}M & 68.92G & $\times$1.34 \\
\midrule
SCEdit/k=1 & 73.18 & 28.82M & \textbf{34.78}G & \textbf{$\times$0.87} \\
SCEdit/k=3 & \textbf{71.78} & 99.11M & 35.28G & \textbf{$\times$0.87} \\
\bottomrule
\end{tabular}
% \vspace{-2mm}
\end{table}
%%%%%%%%%%%%%%%%%%%%%%%%%%%%%%%%%%%%%%%%

%%%%%%%%%%%%%%%%%%%%%%%%%%%%%%%%%%%%%%%%
\begin{figure*}[!t]
    \includegraphics[width=1.0\linewidth]{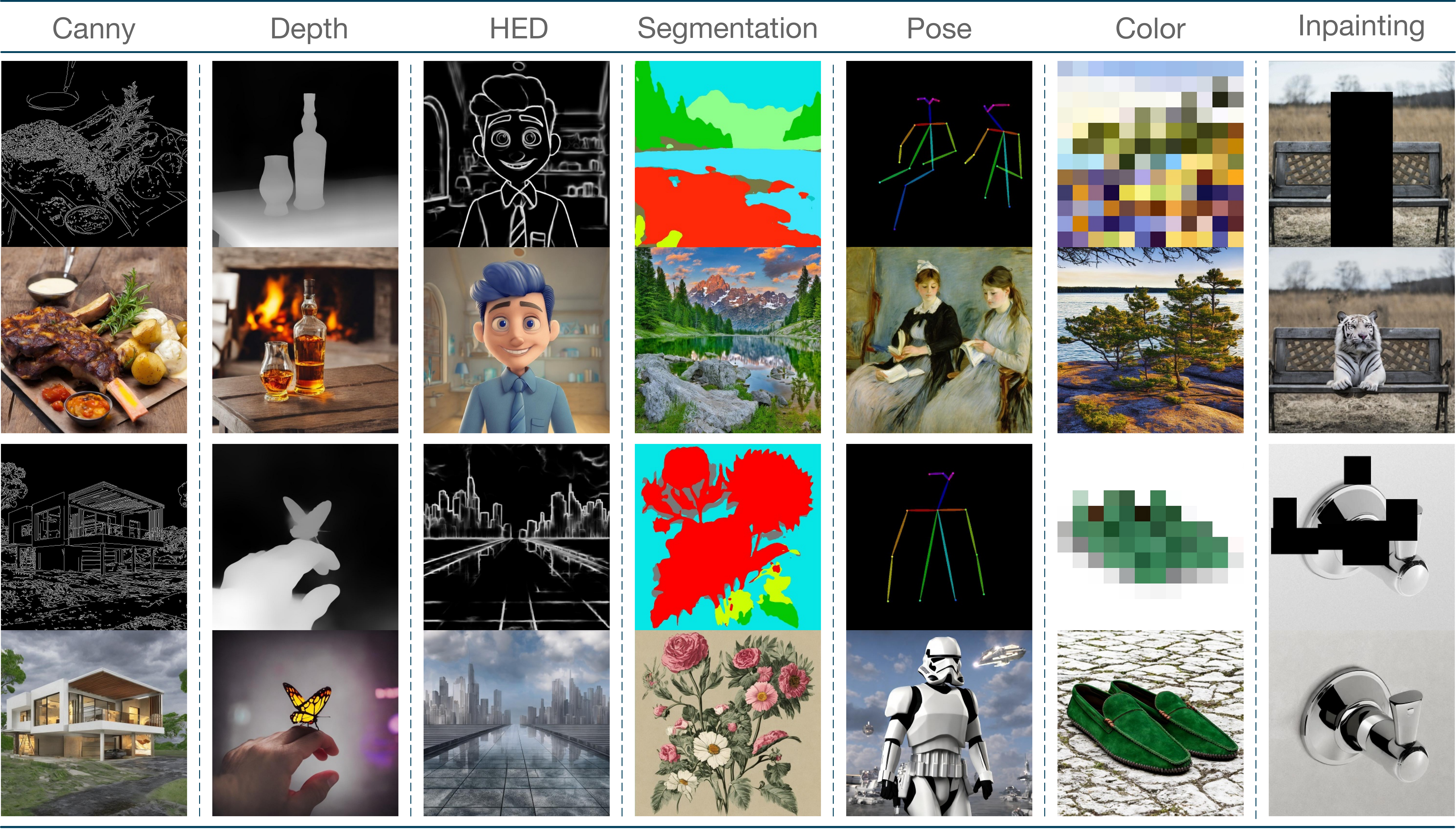}
    \caption{\textbf{Controllable image synthesis with various conditions.} 
    The odd-numbered rows represent the input conditions, while the even-numbered rows correspond to the generated results.
    SCEdit is capable of generating high-quality images precise to input conditions.}
    \label{figa:control_img_show}
    \vspace{-4pt}
\end{figure*}
%%%%%%%%%%%%%%%%%%%%%%%%%%%%%%%%%%%%%%%%

%%%%%%%%%%%%%%%%%%%%%%%%%%%%%%%%%%%%%%%%
\begin{figure*}[t]
    \includegraphics[width=1.0\linewidth]{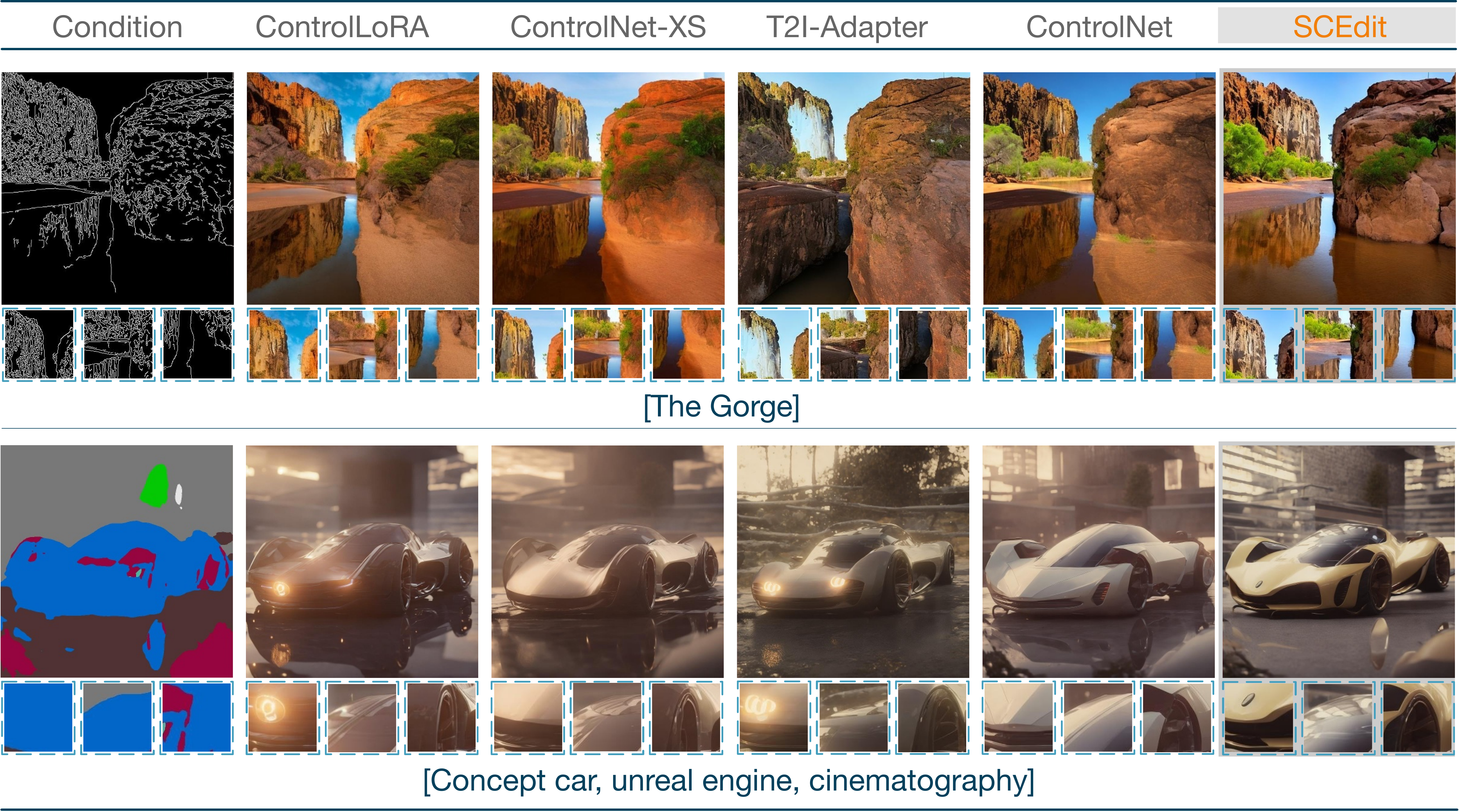}
    \caption{\textbf{Qualitative comparison} to the state-of-the-art controllable image synthesis methods based on the canny edge and semantic segmentation conditions.
    The areas in the boxes underneath are enlarged for detailed comparisons.
    }
    \label{figa:control_img_cmp}
    \vspace{-5pt}
\end{figure*}
%%%%%%%%%%%%%%%%%%%%%%%%%%%%%%%%%%%%%%%%

\subsection{Controllable image synthesis}

\noindent
\textbf{Performance and efficiency comparison.}
Extensive experiments are conducted to evaluate the effectiveness of our proposed SCEdit using CSC-Tuner on various conditional generation tasks, \textit{e.g.}, canny edge, depth, hed boundary, semantic segmentation, pose keypoint, color, masked image, and so on. 
As illustrated in ~\cref{figa:control_img_show}, our method shows superior performance and precise control ability under diverse scenarios and styles control, including real scenes, imagined scenarios, and artistic styles.

We also compare our approach with the state-of-the-art methods that generate images from different conditions.
From a qualitative perspective, we present the generated results based on the canny edge map  and semantic segmentation conditions with different methods and zoom-in on the detailed parts of the image in ~\cref{figa:control_img_cmp}.
It can be observed that our method achieves better quality in terms of fidelity and realism, such as the preservation of the reflection on the lake surface in the first row and the texture information of the car in the second row.
From a quantitative perspective, as seen in ~\cref{taba:control_effi_cmp}, our method utilizes only 7.9\% of the number of parameters in ControlNet, resulting in a 30\% reduction in memory consumption, and accelerates the training process while also yielding a lower FID score.

%%%%%%%%%%%%%%%%%%%%%%%%%%%%%%%%%%%%%%%%
\begin{figure*}[t]
    \centering
    \subfloat[Using the same canny edge map and textual prompt in combination with different color maps results in the depiction of a mountain across various seasons.]{
        \includegraphics[width=1.0\linewidth]{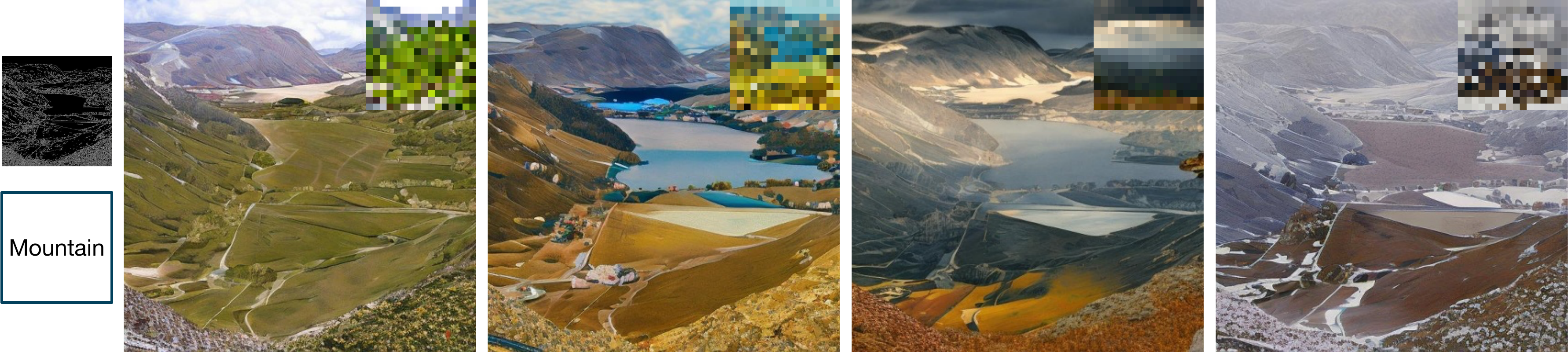}
        \label{figa:dualcontrol}
    }
    \vspace{3mm}
    \subfloat[Interpolations within the canny edge map and color maps.]{
        \includegraphics[width=1.0\linewidth]{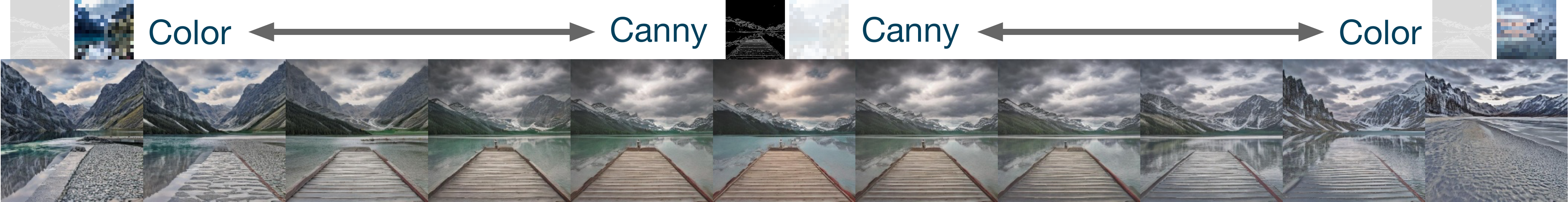}
        \label{figa:interpolation}
    }
    \vspace{-1mm}
    \caption{\textbf{Composable generation.} Combinations of multi-conditions provide more compositional possibilities.}
    \label{figa:multicontrol}
    \vspace{-4mm}
\end{figure*}
%%%%%%%%%%%%%%%%%%%%%%%%%%%%%%%%%%%%%%%%

\noindent
\textbf{Composable generation.}
In addition to generating control based on a single condition, we also support the input of multi-conditions simultaneously. 
In ~\cref{figa:multicontrol}, we demonstrate the joint application of separately trained canny edge and color map models and present the results on unpaired condition data for training-free scene-level image translation. 
By anchoring the controlled subject to a canny edge map and text prompt, and incorporating various color maps, we produce a distinctive seasonal transition effect, as illustrated in ~\cref{figa:dualcontrol}.
Additionally, by traversing the embedding space representations of CSC-Tuner between the two conditions, we can blend them to achieve variations. In ~\cref{figa:interpolation}, SCEdit further provides precise control capabilities, enabling different interpolation effects through balancing among multiple elements.

%%%%%%%%%%%%%%%%%%%%%%%%%%%%%%%%%%%%%%%%
\begin{figure}[t]
    \includegraphics[width=1.0\linewidth]{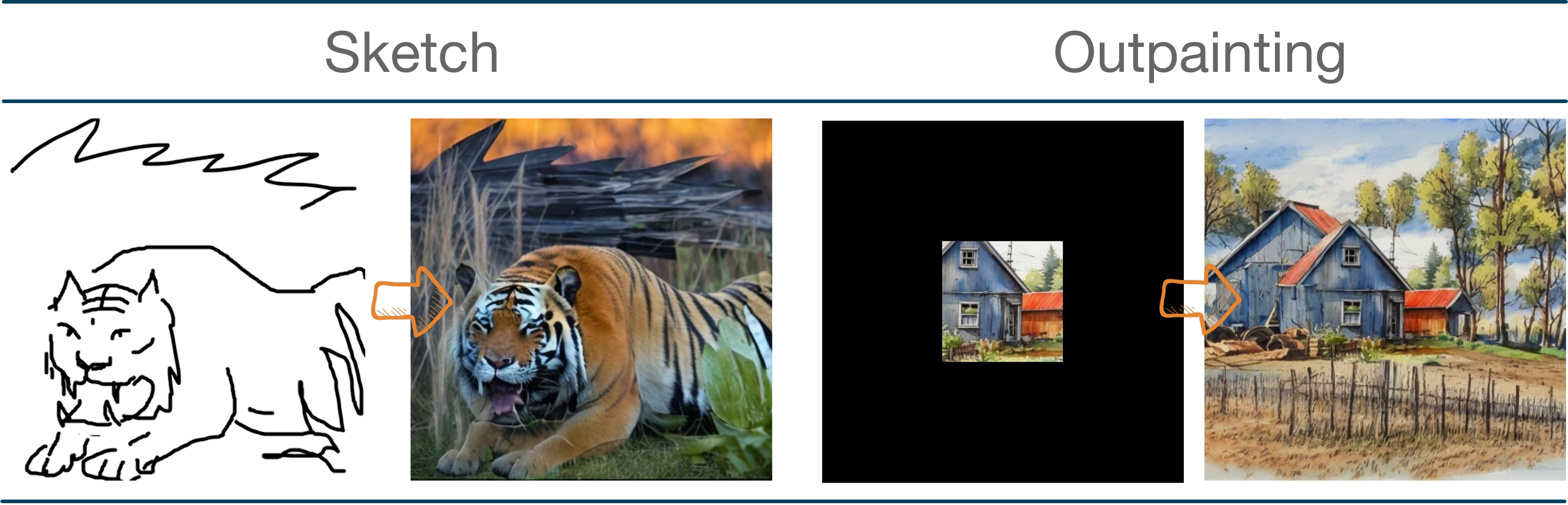}
    \caption{\textbf{Control generalization.} Learned edge condition model is capable of sketch-to-image task and inpainting condition model can control outpainting generation.}
    \label{figa:control_generalization}
    \vspace{-5mm}
\end{figure}
%%%%%%%%%%%%%%%%%%%%%%%%%%%%%%%%%%%%%%%%

\noindent
\textbf{Controllable generalization.}
We find that additional controllable generation tasks could be attempted on models that had already been trained under specific conditions. 
In ~\cref{figa:control_generalization}, we achieve impressive results using the edge-based model for the sketch-to-image task and the inpainting-based model for the outpainting task, although there is no adaptation based on non-real conditions such as hand-drawn sketches for the former, and no specialized adjustment of mask generation patterns for outpainting for the latter.

\section{Conclusion}
\label{sec:conclusion}
We propose SCEdit as an efficient and controllable method for image diffusion generation. 
We introduce SC-Tuner to edit skip connections and extends it to CSC-Tuner, enabling a diverse range of conditional inputs. 
Our method achieves high efficiency by exclusively performing backward propagation in the decoupled U-Net decoder. 
As a lightweight and plug-and-play module, SCEdit supports arbitrary single- and multi-condition generation and demonstrates remarkable superiority in terms of performance.

\clearpage
\clearpage

{
    \small
    \bibliographystyle{ieeenat_fullname}
    \bibliography{main}
}

\clearpage
\appendix

\noindent
In the appendix, we provide more implementation details (\cref{supa:impl}) including the dataset, architecture design, and hyperparameters used in training and inference. 
Then, we demonstrate the ablation experiments (\cref{supa:abl}) with SC-Tuner and CSC-Tuner on different tasks. 
Furthermore, we showcase additional comparisons with existing methods and qualitative results (\cref{supa:addit}). 

\section{Implementation details}
\label{supa:impl}

\subsection{Dataset description}
In this work, we consider three datasets for our experiments: COCO Dataset~\cite{coco}, Customized Style Dataset~\cite{customstyle}, and LAION Dataset~\cite{laion5b}. For the text-to-image generation setting, we utilize the well-known COCO2017 Captions, which consists of 118,287 training images and 591,753 captions for efficient fine-tuning, and Customized Style, which contains 30 training images of different styles for few-shot fine-tuning. We use LAION Dataset for the controllable image synthesis setting. The three datasets are illustrated in ~\cref{taba:dataset}.

\begin{table*}[bp]
\setlength\tabcolsep{7.3pt}
\caption{The summary of the datasets for the experiments.}
% \vskip 0.15in
\centering
\begin{tabular}{l|l|l|ll|ll}
\toprule
\multirow{2}{*}{\textbf{Dataset}} & \multirow{2}{*}{\textbf{\#Description}} & \multirow{2}{*}{\textbf{\#Task}} & \multicolumn{2}{c|}{\textbf{\#Train}} & \multicolumn{2}{c}{\textbf{\#Test}} \\
 &  & & image & prompt & image & prompt \\

\midrule
\multicolumn{7}{l}{\textit{Common Objects in Context (COCO)}} \\
COCO2017 Captions~\cite{coco} & common objects & text-to-image & 118,287 & 591,753 & 5,000 & 25,014 \\

\midrule
\multicolumn{7}{l}{\textit{Customized Style Dataset}} \\
3D~\cite{customstyle} & 3D style & text-to-image (few-shot) & 30 & 30 & - & - \\
Anime~\cite{customstyle} & ainme style & text-to-image (few-shot) & 30 & 30 & - & - \\
Flatillustration~\cite{customstyle} & flatillustration style & text-to-image (few-shot) & 30 & 30 & - & - \\
Oilpainting~\cite{customstyle} & oilpainting style & text-to-image (few-shot) & 30 & 30 & - & - \\
Sketch~\cite{customstyle} & sketch style & text-to-image (few-shot) & 30 & 30 & - & - \\
Watercolor~\cite{customstyle} & watercolor style & text-to-image (few-shot) & 30 & 30 & - & - \\

\midrule
\multicolumn{7}{l}{\textit{Large-scale Artificial Intelligence Open Network (LAION)}} \\
LAION-ART~\cite{laion5b} & filtered version & controllable generation & 624,558 & 624,558 & - & - \\

\bottomrule
\end{tabular}
\label{taba:dataset}
% \vskip -0.1in
\end{table*}

\subsection{Hyperparameters}
We provide an overview of the hyperparameters for all trained models, divided by the task in ~\cref{taba:hyper}.

\begin{table*}[ht]
\caption{The summary of the training and inference settings for the experiments.}
% \vskip 0.15in
\setlength\tabcolsep{14.8pt}
\centering
\begin{tabular}{l|ccc}
\toprule
\multirow{2}{*}{\textbf{Config}} & \multicolumn{3}{c}{\textbf{\#Task}} \\
 & Text-to-image & Text-to-image (few-shot) & Controllable Generation \\
 
\midrule
Dataset & COCO~\cite{coco} & Customized Style~\cite{customstyle} & LAION-ART (Filtered)~\cite{laion5b} \\
Batch size & 32 & 8 & 64 \\
Optimizer & AdamW~\cite{adamw} & AdamW~\cite{adamw} & AdamW~\cite{adamw} \\
Weight decay & 0.01 & 0.01 & 0.01 \\
Learning rate & 0.00005 & 0.00005 & 0.00005 \\
Learning rate schedule & Constant & Constant & Constant \\
Training steps & 100000 & 1500 & 100000 \\
Data preprocess & Resize, CenterCrop & Resize, CenterCrop & Resize, CenterCrop \\
Resolution & 512$\times$512 & 512$\times$512 & 512$\times$512 \\
Pre-trained & SD v1.5~\cite{sd15} & SD v1.5~\cite{sd15} & SD v2.1~\cite{sd21} \\

\midrule
Sampler & DDIM~\cite{ddim} & DDIM~\cite{ddim} & DDIM~\cite{ddim} \\
Sample steps & 50 & 50 & 50 \\
Guide scale & 3.0 & 7.5 & 7.5 \\

\midrule
Device & A100$\times$8 & A100$\times$1 & A100$\times$16 \\
Training strategy & AMP / Float16 & AMP / Float16 & AMP / Float16 \\
Library & SWIFT~\cite{swift} & SWIFT~\cite{swift} & SWIFT~\cite{swift} \\

\bottomrule
\end{tabular}
\label{taba:hyper}
% \vskip -0.1in
\vspace{3pt}
\end{table*}

\subsection{Architectures design}
In the SCEdit framework, the central strategy involves editing the skip connections, which gives rise to two architectures: SC-Tuner for text-to-image generation and CSC-Tuner for controllable generation. These architectures are straightforward to implement and can be easily transferred to other similarly designed modules. In ~\cref{alg:code}, we provide the forward function implementation of SCEdit written in PyTorch-like style.

%%%%%%%%%%%%%%%%%%%%%%%%%%%%%%%%%%%%%%%%
\begin{algorithm*}[ht]
\caption{Implementation of SCEdit in PyTorch-like style.}\label{alg:code}
% \vspace{50mm}
\definecolor{codeblue}{rgb}{0.25,0.5,0.5}
\definecolor{colorred}{RGB}{197, 49, 124}
\lstset{
  backgroundcolor=\color{white},
  basicstyle=\fontsize{8.8pt}{8.8pt}\ttfamily\selectfont,
  columns=fullflexible,
  breaklines=true,
  captionpos=b,
  commentstyle=\fontsize{7.2pt}{7.2pt}\color{codeblue},
  keywordstyle=\fontsize{7.2pt}{7.2pt}\color{colorred},
}
\begin{minipage}[ht]{0.49\textwidth}
\begin{lstlisting}[language=python]
# SC-Tuner
def forward(self, x, t=None, cond=dict()):
    ...
    # input_blocks
    hs = []
    for i, blk in enumerate(self.in_blks):
        h = blk(h, emb, context)
        hs.append(h)
        
    # middle_block
    h = self.mid_blk(h, emb, context)
    
    # output_blocks
    for i, blk in enumerate(self.out_blks):
        skip_h = self.tuners[i](hs.pop())
        h = torch.cat([h, skip_h], dim=1)
        h = blk(h, emb, context)
\end{lstlisting}
\end{minipage}%
\hfill
\vrule{}
\hfill
\begin{minipage}[ht]{0.49\textwidth}
\begin{lstlisting}[language=python]
# Single CSC-Tuner
def forward(self, x, t=None, cond=dict()):
    ...
    # Dense Conv for conditions
    guid_hs = []
    guid_hint = self.in_hint_blks(hint, emb, context)
    for i, blk in enumerate(self.hint_blks):
        guid_hint = blk(guid_hint, emb, context)
        guid_hs.append(guid_hint)
    ...
    # output_blocks
    for i, blk in enumerate(self.out_blks):
        skip_h = self.tuners[i](hs.pop() + self.scale * guid_hs[::-1][i])
        h = torch.cat([h, skip_h], dim=1)
        h = blk(h, emb, context)
\end{lstlisting}
\end{minipage}
\end{algorithm*}
%%%%%%%%%%%%%%%%%%%%%%%%%%%%%%%%%%%%%%%%

\subsection{Conditions for generation}

We generally follow the implementations of condition extraction from ControlNet~\cite{controlnet} and T2I-Adapter~\cite{t2i-adapter}, with details as follows:
\begin{itemize}
\item \textbf{Canny Edge Map.} We employ canny edge detector~\cite{canny}, utilizing random thresholds during training and fixed thresholds with a low value of 100 and a high value of 200 during inference. The sample images are presented in ~\cref{supa:control_img_canny}.
\item \textbf{Depth Map.} We use MiDaS depth estimation~\cite{depth} with default settings. The sample images are shown in ~\cref{supa:control_img_depth}.
\item \textbf{HED Boundary Map.} We use HED boundary detection~\cite{hed} with default settings. The sample images are illustrated in ~\cref{supa:control_img_hed}. 
\item \textbf{Semantic Segmentation Map.} We employ the UniFormer~\cite{uniformer} semantic segmentation model, which was trained on the ADE20K~\cite{ade20k} dataset. The sample images can be seen in ~\cref{supa:control_img_seg}. 
\item \textbf{Pose Keypoint.} We employ OpenPose~\cite{openpose} as the human pose estimation model and visualize its prediction as conditions. The sample images are showcased in ~\cref{supa:control_img_pose}.
\item \textbf{Color Map.} We preserve the spatial hierarchical color information through a process of 64$\times$ downsampling of the image, subsequently followed by an upsampling to its original dimensions. The sample images are demonstrated in ~\cref{supa:control_img_color}. 
\item \textbf{Inpainting.} We employ the mask generation strategy from LaMa~\cite{lama} for conditional generation on the inpainting task. The sample images are demonstrated in ~\cref{supa:control_img_inpainting,supa:control_img_outpainting}. 
\end{itemize}

\noindent
For all the aforementioned conditions, we utilize the same training dataset (LAION-ART~\cite{laion5b}) and hyperparameters across the tasks. 
The exception is the pose-conditional task, for which we exclusively utilize a subset of images containing human poses, amounting to a total of 162,338 instances.
Additionally, for the inpainting task, we follow the common approach of using both masks and cutouts as combined conditional inputs.

\begin{table*}[ht]
% \vskip 0.15in
\centering
\caption{\textbf{SC-Tuner ablation} experiments of efficient fine-tuning task on COCO2017. Default settings are marked in gray.}
    \begin{subtable}{.31\textwidth}
    \centering
    \caption{Ablation on downscaling ratio of dimensions.}
    \label{taba:ft_abl_dim}
    \setlength\tabcolsep{5pt}
    \begin{tabular}{llll}
    \toprule
    Ratio & FID & Params & Mem. \\
    \midrule
    \rowcolor{tabhighlight} $\times$1 & 13.82 & 19.68M & 29.02G \\
    $\times$5 & 13.92 & 3.94M & 28.29G \\
    $\times$10 & 13.99 & 1.98M & 28.06G \\
    \bottomrule
    \end{tabular}
    \end{subtable}
\hfill
    \begin{subtable}{.33\textwidth}
    \centering
    \caption{Ablation on skip connection (SC) layers.}
    \label{taba:ft_abl_idx}
    \setlength\tabcolsep{3pt}
    \begin{tabular}{llll}
    \toprule
    SC Indexes & FID & Params & Mem. \\
    \midrule
    \{0,11\} & 14.45 & 3.48M & 28.11G \\
    \{0,3,6,9,11\} & 13.96 & 7.79M & 28.56G \\
    \rowcolor{tabhighlight} \{1,2, ..., 12\} & 13.82 & 19.68M & 29.02G \\
    \bottomrule
    \end{tabular}
    \end{subtable}
\hfill
    \begin{subtable}{.31\textwidth}
    \centering
    \caption{Ablation on tuner submodules.}
    \label{taba:ft_abl_module}
    \setlength\tabcolsep{2.5pt}
    \begin{tabular}{llll}
    \toprule
    Module & FID & Params & Mem. \\
    \midrule
    \rowcolor{tabhighlight} Linear & 13.82 & 19.68M & 29.02G \\
    Conv & 13.88 & 22.13M & 28.65G \\
    ResPrefix~\cite{restuning} & 14.38 & 21.64M & 30.54G \\
    \bottomrule
    \end{tabular}
    \end{subtable}
\label{taba:ft_abl}
\vspace{5pt}
\end{table*}

\begin{table*}[ht]
% \vskip 0.1in
\centering
\caption{\textbf{CSC-Tuner ablation} experiments of controllable generation task on LAION dataset. Default settings are marked in gray.}
    \begin{subtable}{.3\textwidth}
    \centering
    \caption{Ablation on convolution kernel size.}
    \label{taba:ctr_abl_kernel}
    \setlength\tabcolsep{5pt}
    \begin{tabular}{llll}
    \toprule    
    Kernel & FID & Params & Mem. \\
    \midrule
    \rowcolor{tabhighlight} 1 & 73.18 & 28.82M & 34.78G \\
    3 & 71.78 & 99.11M & 35.28G \\
    \bottomrule
    \end{tabular}
    \end{subtable}
\hfill
    \begin{subtable}{.33\textwidth}
    \centering
    \caption{Ablation on skip connection (SC) layers.}
    \label{taba:ctr_abl_idx}
    \setlength\tabcolsep{2pt}
    \begin{tabular}{llll}
    \toprule
    SC Indexes & FID & Params & Mem. \\
    \midrule
    \{0,3,4,6,7,9,11\} & 85.42 & 17.14M & 34.48G \\
    \rowcolor{tabhighlight} \{1,2,3, ..., 12\} & 73.18 & 28.82M & 34.78G \\
    \bottomrule
    \end{tabular}
    \end{subtable}
\hfill
    \begin{subtable}{.3\textwidth}
    \centering
    \caption{Ablation on tuner submodules.}
    \label{taba:ctr_abl_module}
    \setlength\tabcolsep{2.5pt}
    \begin{tabular}{llll}
    \toprule
    Module & FID & Params & Mem. \\
    \midrule
    \rowcolor{tabhighlight} Single Conv & 73.18 & 28.82M & 34.78G \\
    Dual Conv & 70.54 & 37.82M & 35.31G \\
    \bottomrule
    \end{tabular}
    \end{subtable}
\label{taba:ctr_abl}
\end{table*}

%%%%%%%%%%%%%%%%%%%%%%%%%%%%%%%%%%%%%%%%
\begin{figure*}[htb]
    % \vspace{-10mm}
    % \setlength{\abovecaptionskip}{0pt}
    % \setlength{\belowcaptionskip}{-4pt}
    \includegraphics[width=1.0\linewidth]{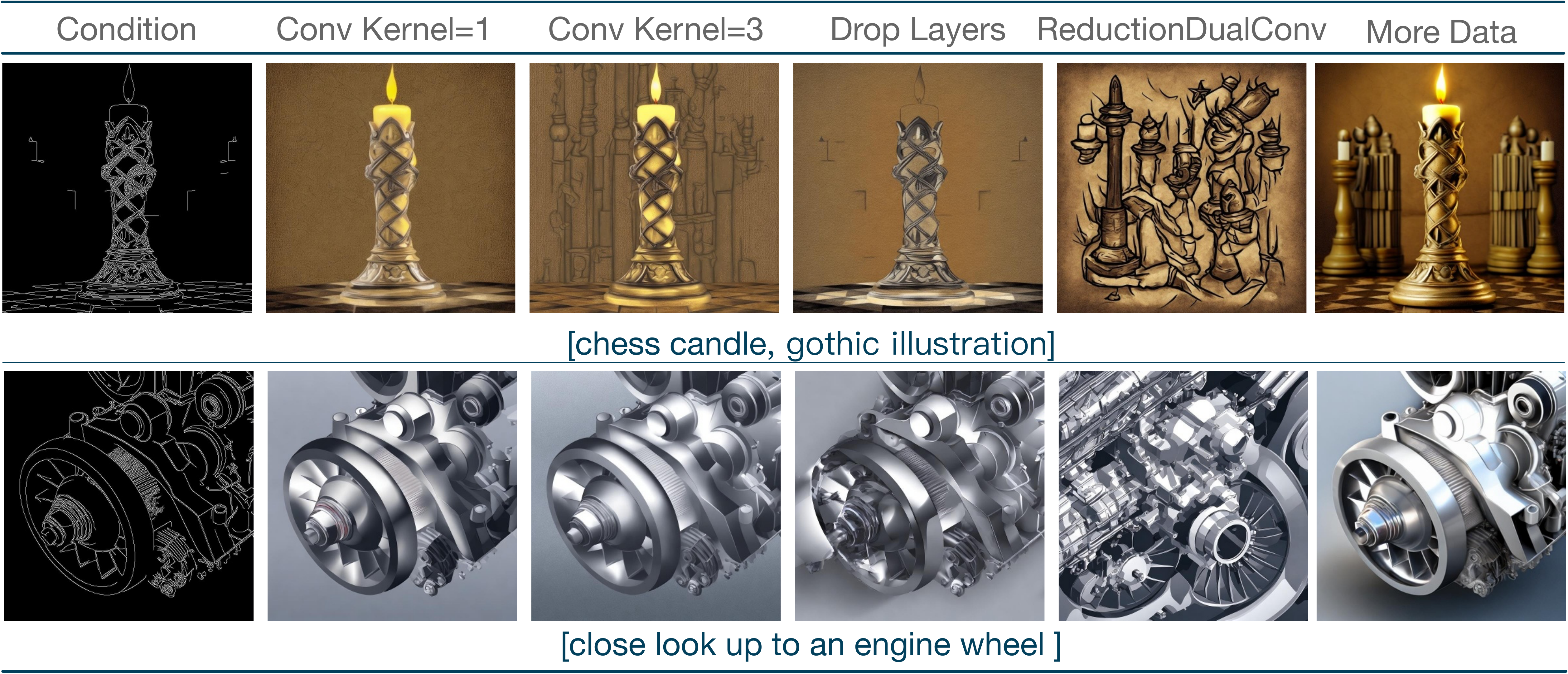}
    \caption{\textbf{Qualitative comparison} on various CSC-Tuner structure designs.}
    \label{supa:control_img_abl}
    \vspace{3mm}
\end{figure*}
%%%%%%%%%%%%%%%%%%%%%%%%%%%%%%%%%%%%%%%%

\section{Ablation studies}
\label{supa:abl}

\subsection{SC-Tuner structure}

We ablate our SC-Tuner using the default setting in ~\cref{taba:ft_abl}. 
It is evident that our method allows for flexible design, including the intermediate dimensions of tuners, the number of utilized skip connection layers, and the selection of submodules.

In ~\cref{taba:ft_abl_dim}, we retain the dimensions of the skip connection features as the default intermediate dimensions for the tuner. 
As the dimensions are reduced proportionally, there is a corresponding decrease in the number of parameters. Despite this reduction, the decline in memory consumption is not substantial, and the FID~\cite{fid} fails to show an improvement compared to the default setting. Similarly, in ~\cref{taba:ft_abl_idx}, a performance degradation is observed when we reduce the number of skip connection layers by intervals. Our SC-Tuner is designed with the flexibility to interchange its internal components, allowing for the use of convolution networks or independent residual networks. As demonstrated in ~\cref{taba:ft_abl_module}, even the most elementary components, such as linear layers, can offer certain advantages while maintaining a comparable number of parameters.

% \noindent
% \textbf{Few-shot}
% \clearpage
\subsection{CSC-Tuner structure}

We conducted a series of ablation studies based on the modular design of the CSC-Tuner to evaluate the impact of each component on the overall performance.

From a quantitative perspective, in ~\cref{taba:ctr_abl_kernel}, we can observe that larger convolution kernels of condition encoder, although resulting in an increase in the number of parameters, also contribute to a certain reduction in the FID. 
In ~\cref{taba:ctr_abl_idx}, omitting some of the skip connections results in an increase in the FID.
Subsequently, as shown in ~\cref{taba:ctr_abl_module}, we ablate with altering the internal structure of the tuner by shifting from a single convolution layer to a dual convolution layer with dimension reduction, resulting in improved FID score. 

From a qualitative perspective, we compared the aforementioned experimental setups and also train on a larger dataset (24M) under the default setting. As evident from ~\cref{supa:control_img_abl}, the enlargement of the convolution kernel size expands the receptive field, achieving richer detail in the generated images. Training with more data also benefits from realistic effects. On the other hand, omitting some of the skip connections generally leads to a loss of image content. The dual convolution with dimension reduction exhibits poor control over conditions, underscoring the importance of the channel dimension in generation.

\section{Additional results}
\label{supa:addit}

\subsection{Additional qualitative comparison}

\begin{figure*}[ht]
    \centering
    \subfloat[Comparative results of generation conditioned on canny edge map.]{
        \includegraphics[width=1.0\linewidth]{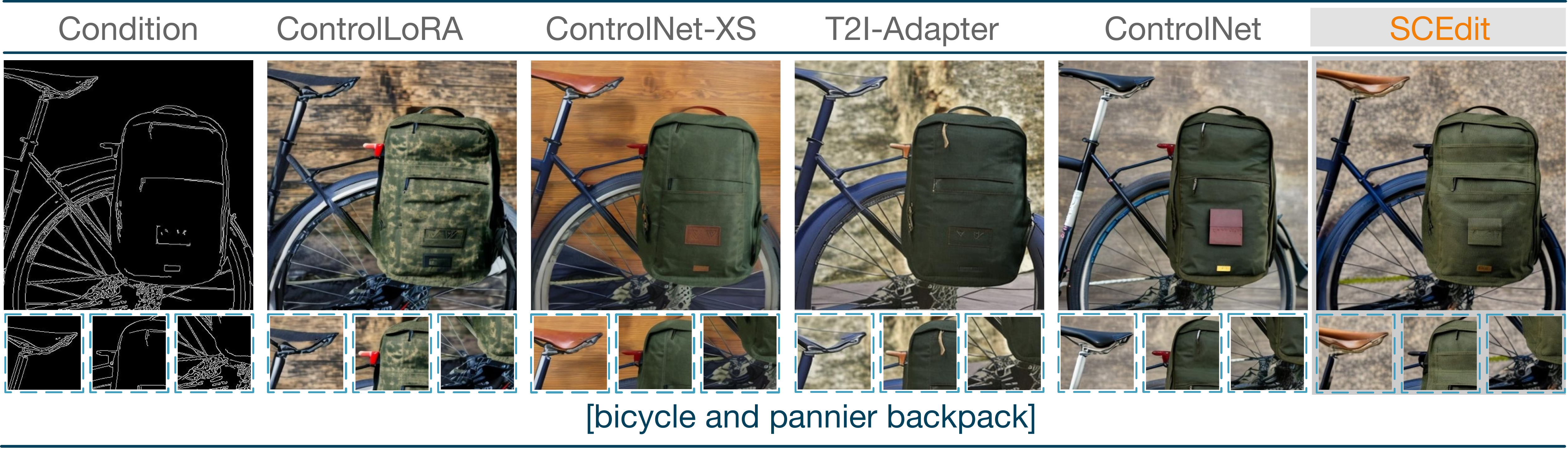}
        \label{supa:control_img_cmp_canny}
    }
    \vspace{5mm}
    \subfloat[Comparative results of generation conditioned on semantic segmentation map.]{
        \includegraphics[width=1.0\linewidth]{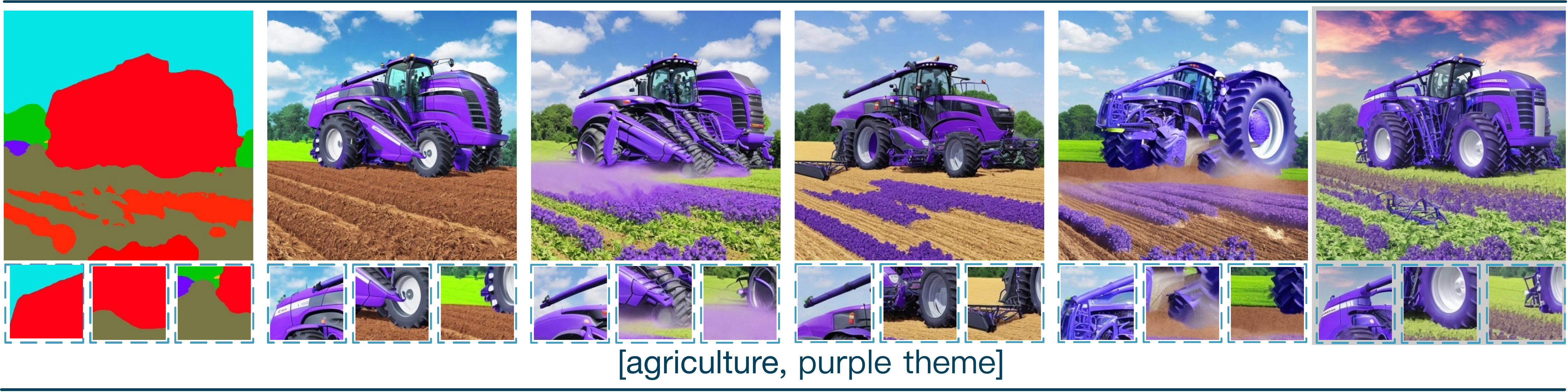}
        \label{supa:control_img_cmp_seg}
    }
    \vspace{5mm}
    \subfloat[Comparative results of generation conditioned on depth map.]{
        \includegraphics[width=1.0\linewidth]{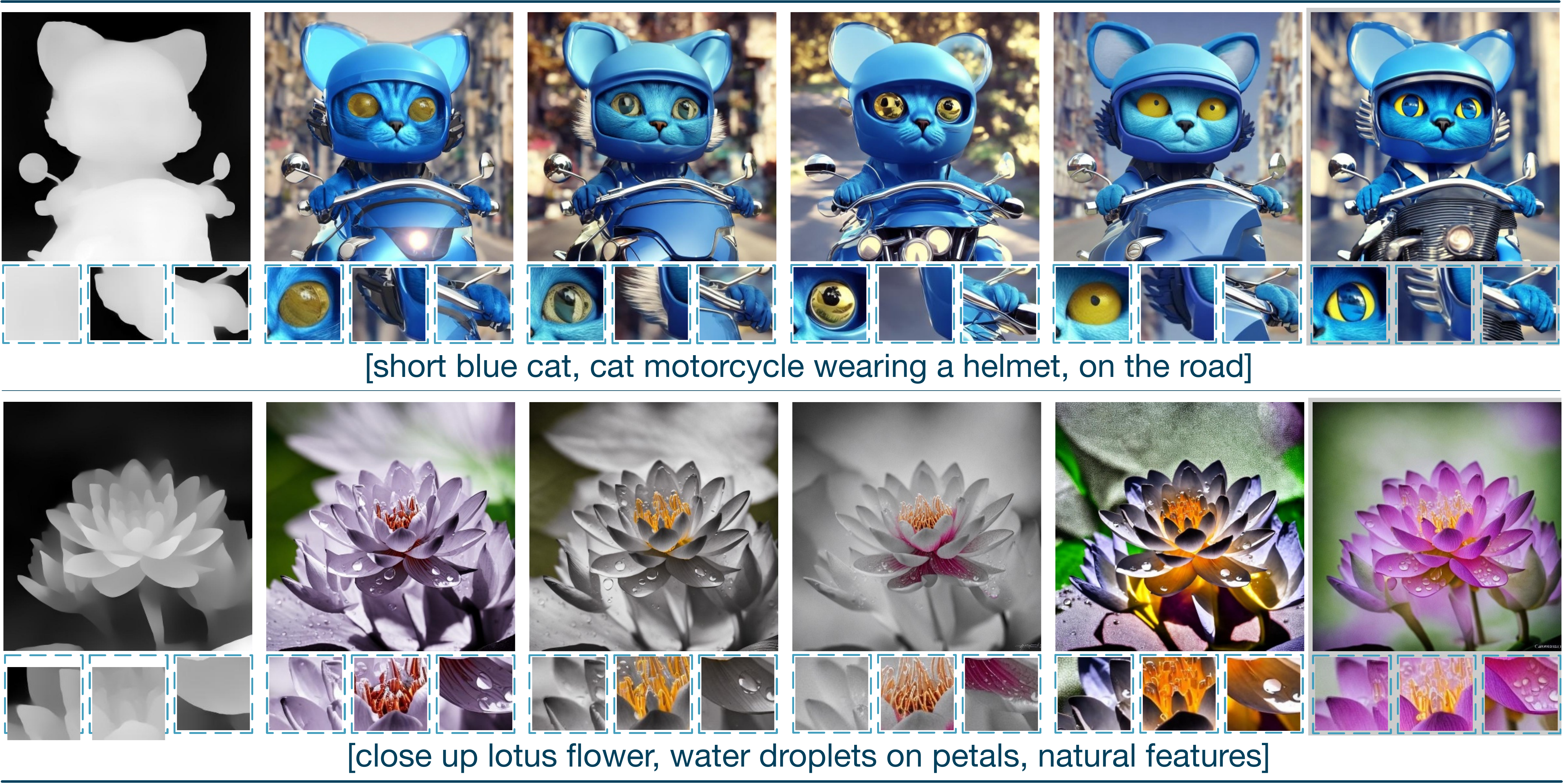}
        \label{supa:control_img_cmp_depth}
    }
    \caption{\textbf{Additional qualitative comparison} on the controllable generation of our approach with other strategies conditioned on canny edge maps, semantic segmentation maps, and depth maps. The areas in the boxes are enlarged for detailed comparisons.}
    \label{supa:control_img_cmp_sup}
\end{figure*}

In~\cref{supa:control_img_cmp_sup}, we present additional qualitative comparison for the controllable generation task, using canny edge maps, depth maps, and
semantic segmentation maps as conditions, including comparisons with methods ControlNet~\cite{controlnet}, T2I-Adapter~\cite{t2i-adapter}, ControlLoRA~\cite{controllora}, and ControlNet-XS~\cite{controlnetxs}.

\subsection{Additional qualitative results}

\begin{figure*}[ht]
    % \vspace{-10mm}
    % \setlength{\abovecaptionskip}{20pt}
    % \setlength{\belowcaptionskip}{-10pt}
    \includegraphics[width=1.0\linewidth]{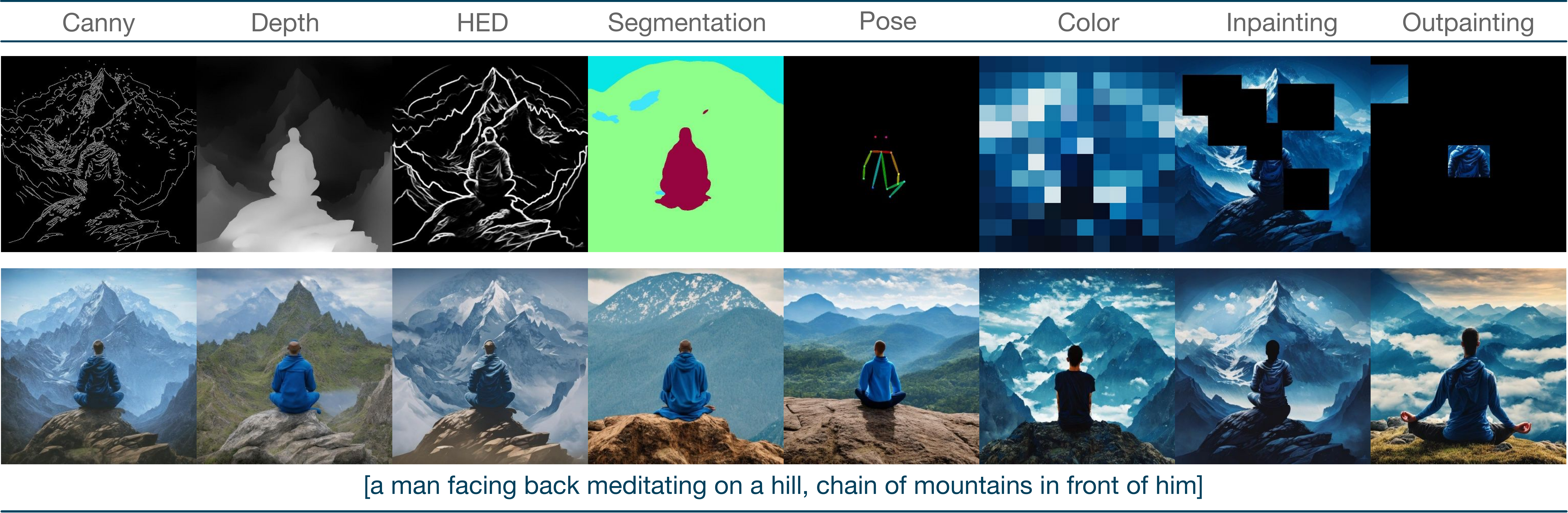}
    \caption{\textbf{Additional qualitative results} on controllable generation using the same original image for different conditions.}
    \label{supa:control_img_all}
    \vspace{-1mm}
\end{figure*}

\begin{figure*}[ht]
    \centering
    \subfloat[Generative results conditioned on canny edge map.]{
        \includegraphics[width=1.0\linewidth]{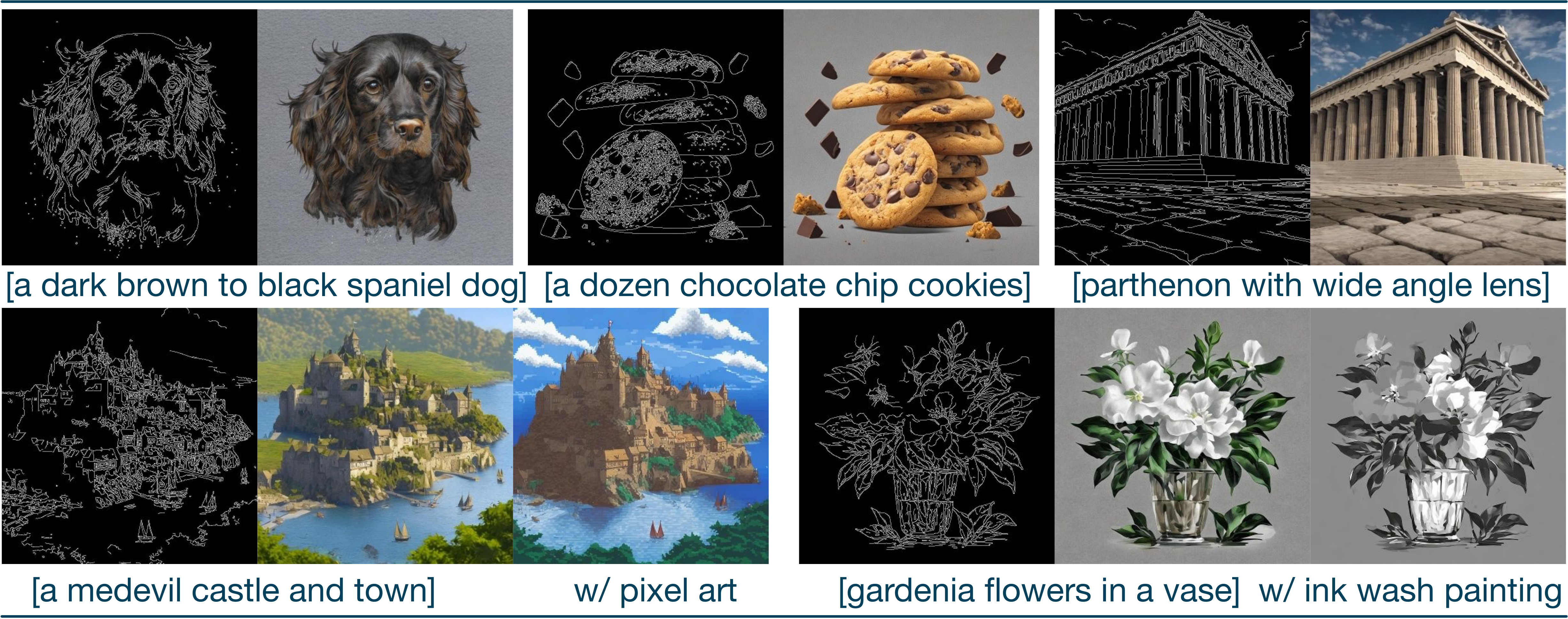}
        \label{supa:control_img_canny}
    }
    \vspace{3mm}
    \subfloat[Generative results conditioned on depth map.]{
        \includegraphics[width=1.0\linewidth]{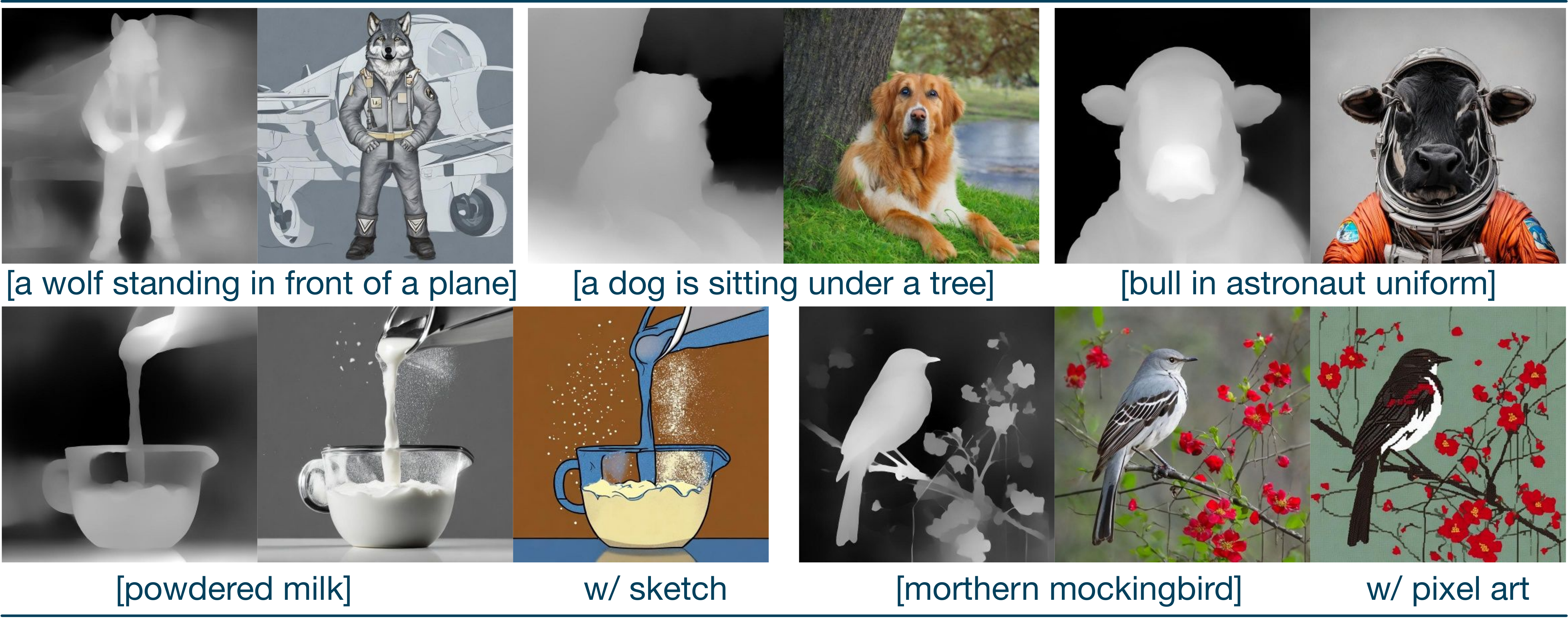}
        \label{supa:control_img_depth}
    }
    \caption{Additional qualitative results on controllable generation using canny edge map and depth map conditions.}
    \label{supa:control_img_show_1}
    \vspace{-5mm}
\end{figure*}

\begin{figure*}[ht]
    \centering
    \subfloat[Generative results conditioned on hed boundary map.]{
        \includegraphics[width=1.0\linewidth]{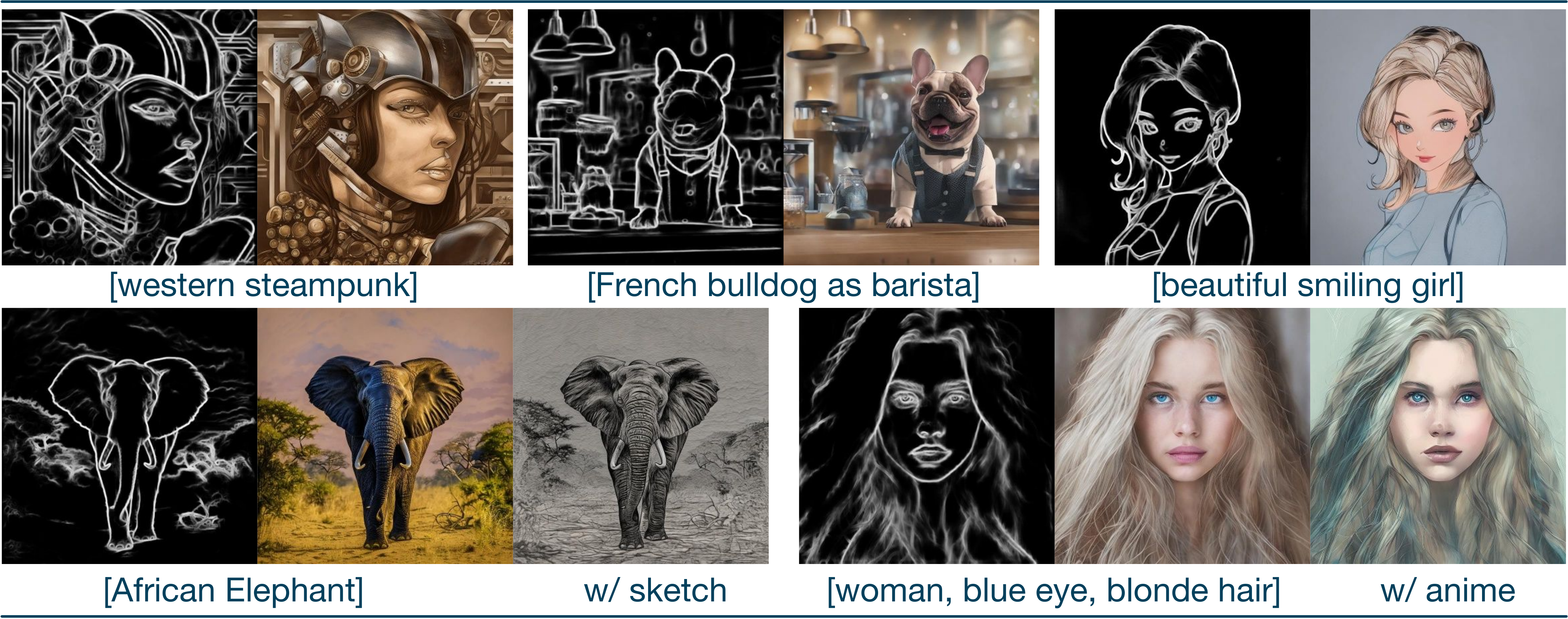}
        \label{supa:control_img_hed}
    }
    \vspace{1mm}
    \subfloat[Generative results conditioned on semantic segmentation map.]{
        \includegraphics[width=1.0\linewidth]{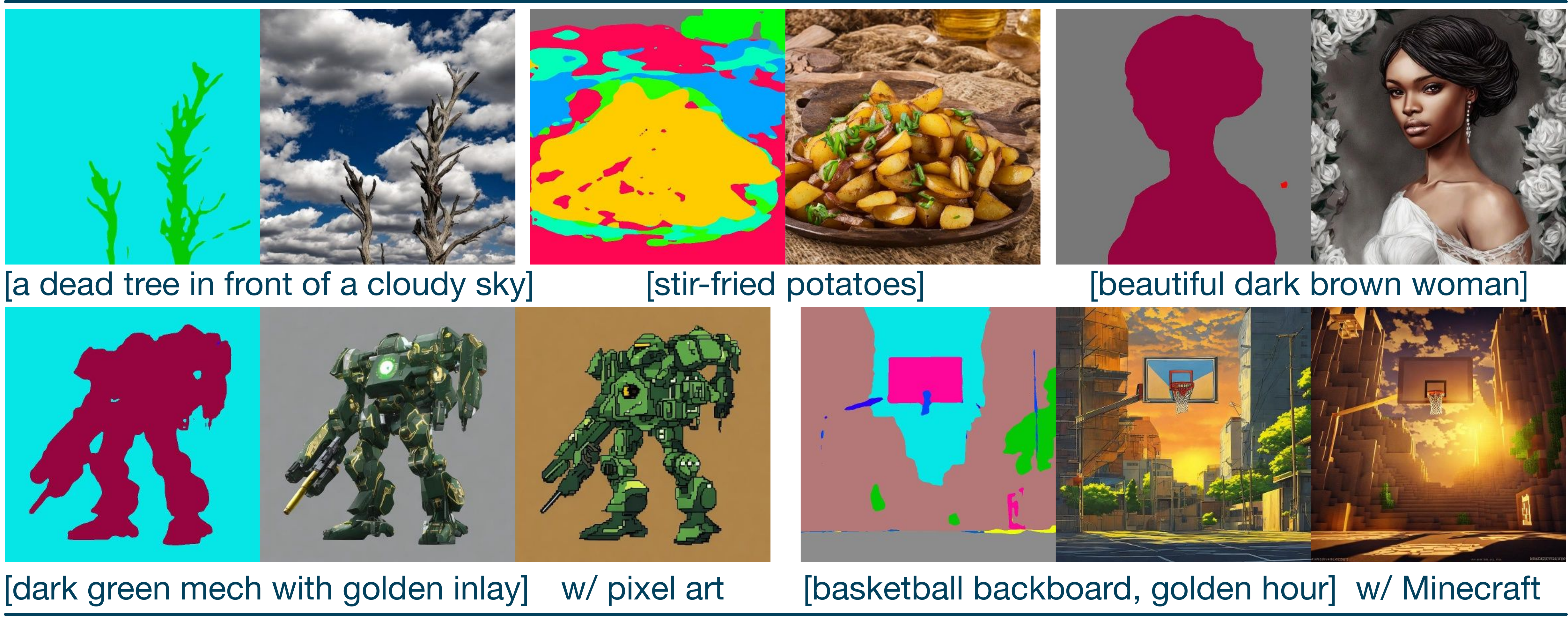}
        \label{supa:control_img_seg}
    }
    \vspace{1mm}
    \subfloat[Generative results conditioned on pose keypoint.]{
        \includegraphics[width=1.0\linewidth]{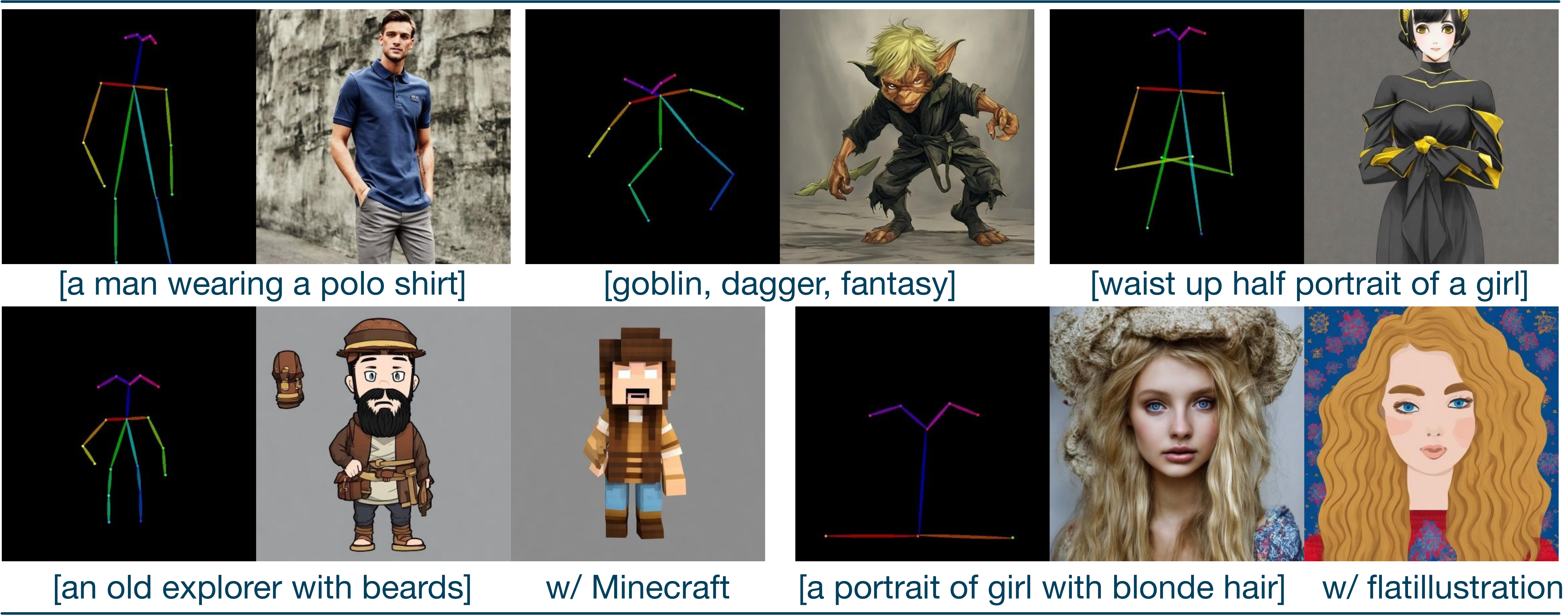}
        \label{supa:control_img_pose}
    }
    \vspace{-1mm}
    \caption{\textbf{Additional qualitative results} on controllable generation using hed boundary map, semantic segmentation map, and pose keypoint conditions.}
    \label{supa:control_img_show_2}
    \vspace{-5mm}
\end{figure*}

\begin{figure*}[ht]
    \centering
    \subfloat[Generative results conditioned on color map.]{
        \includegraphics[width=1.0\linewidth]{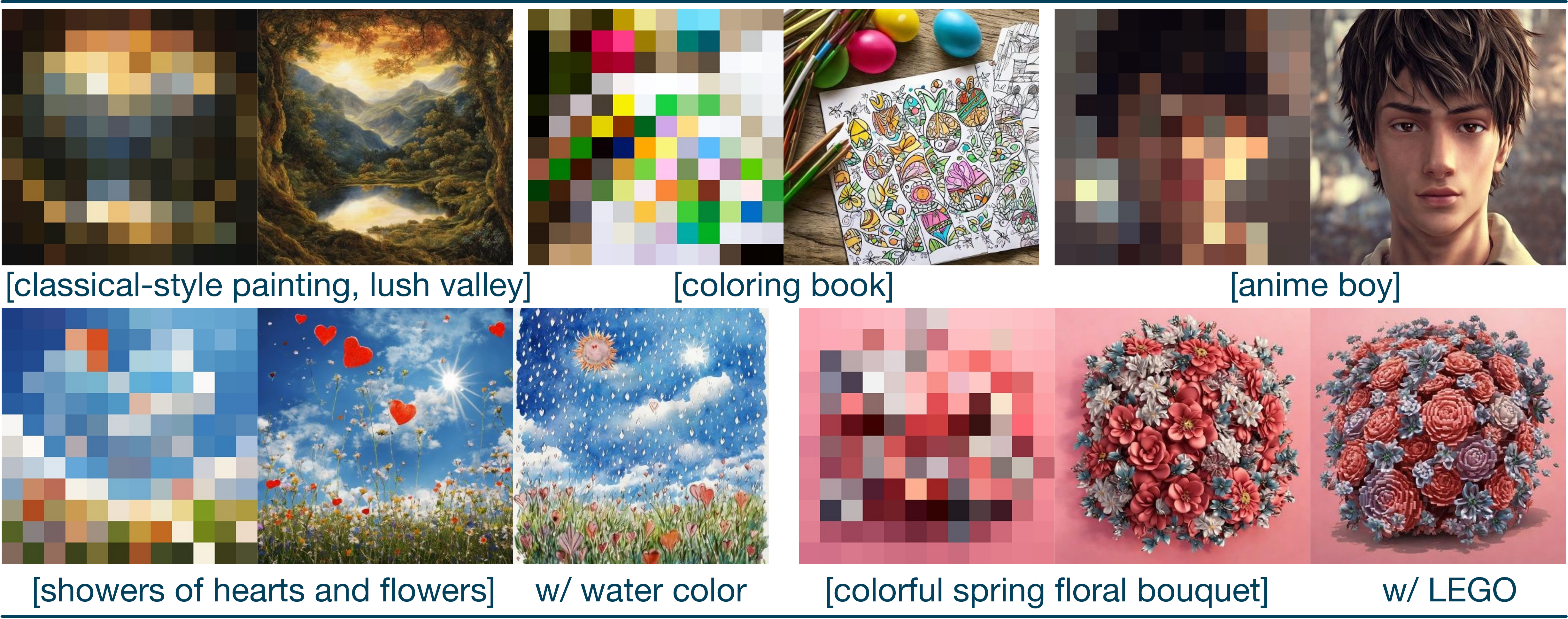}
        \label{supa:control_img_color}
    }
    \vspace{1mm}
    \subfloat[Generative results conditioned on outpainting.]{
        \includegraphics[width=1.0\linewidth]{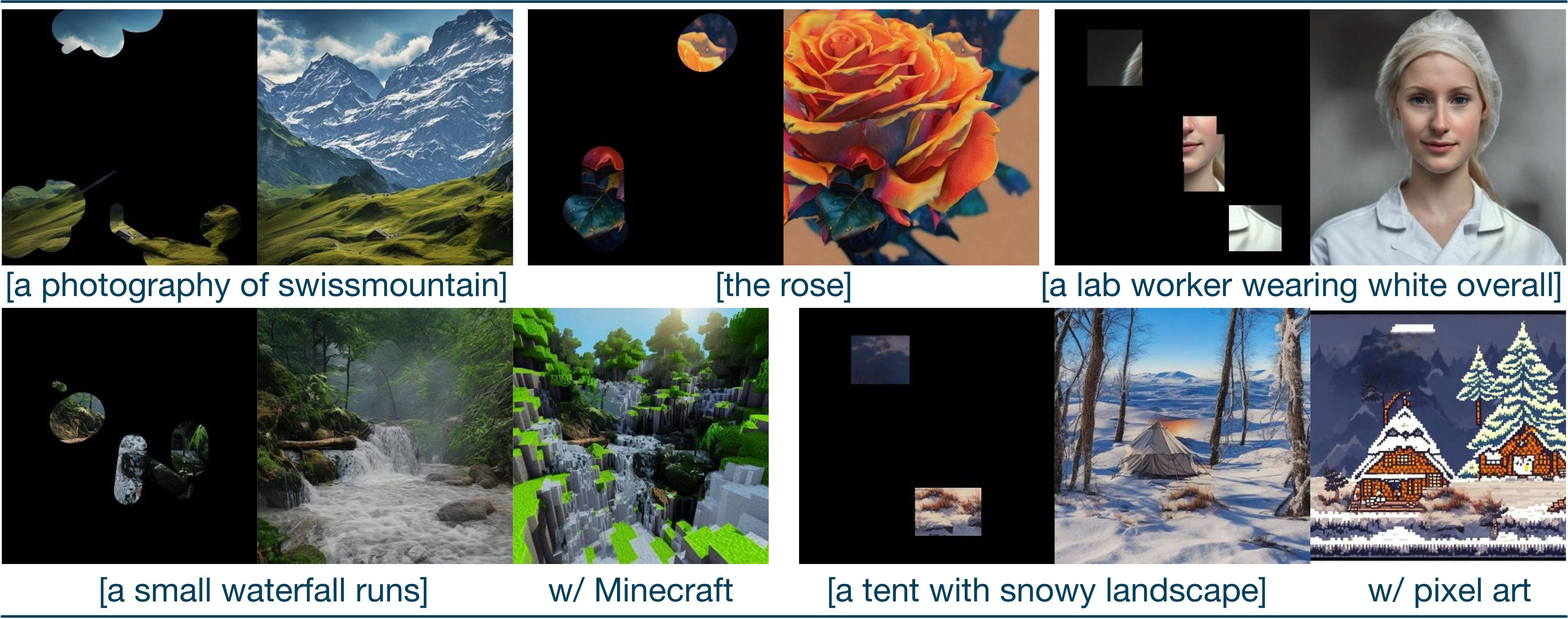}
        \label{supa:control_img_outpainting}
    }
    \vspace{1mm}
    \subfloat[Generative results conditioned on inpainting.]{
        \includegraphics[width=1.0\linewidth]{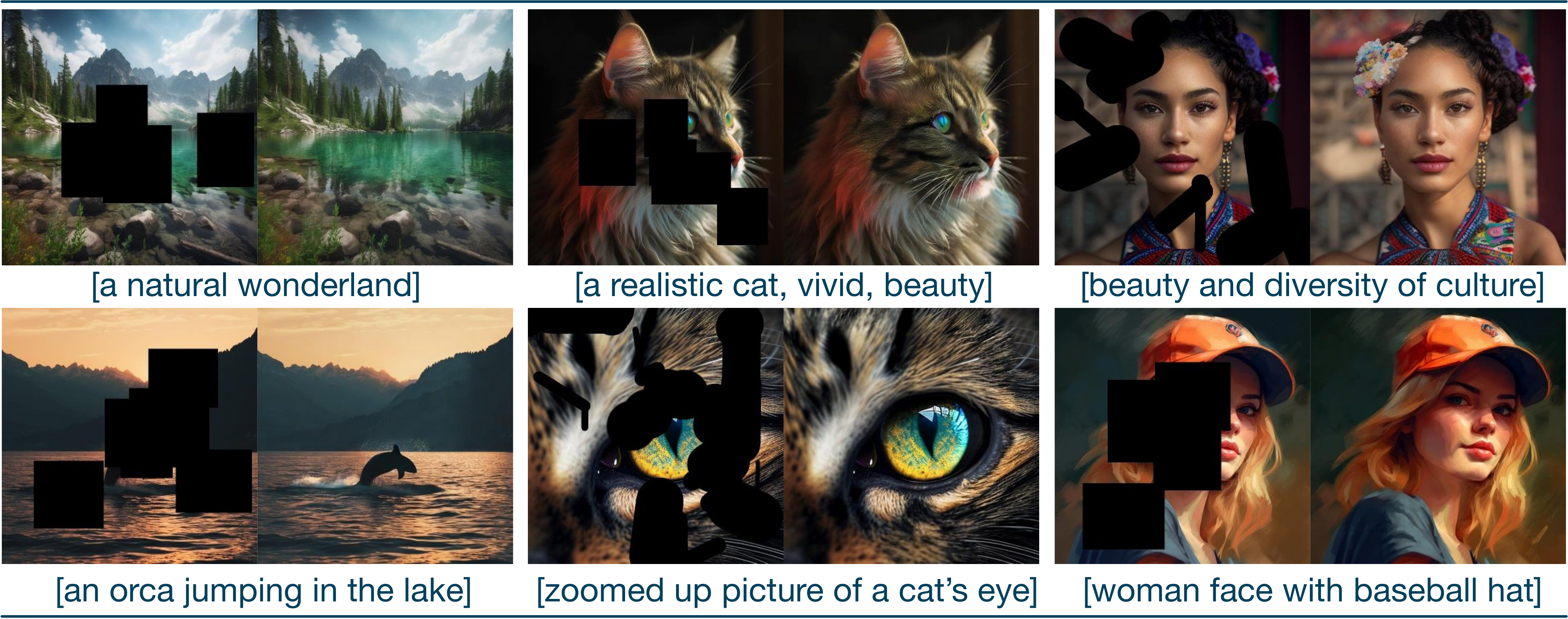}
        \label{supa:control_img_inpainting}
    }
    \vspace{-1mm}
    \caption{\textbf{Additional qualitative results} on controllable generation using color maps, outpainting, and inpainting conditions.}
    \label{supa:control_img_show_3}
    \vspace{-5mm}
\end{figure*}

In~\cref{supa:control_img_all}, we demonstrate the results of generating images by extracting different conditional information from the same image and using it as control conditions.
In~\cref{supa:control_img_show_1}, ~\cref{supa:control_img_show_2}, and ~\cref{supa:control_img_show_3}, we present additional qualitative results for the controllable generation task, with conditions including canny edge map, depth map, hed boundary map, semantic segmentation map, pose keypoint, color map, outpainting, and inpainting.

%%%%%%%%%%%%%%%%%%%%%%%%%%%%%%%%%%%%%%%%
%%%%%%%%%%%%%%%%%%%%%%%%%%%%%%%%%%%%%%%%

%%%%%%%%%%%%%%%%%%%%%%%%%%%%%%%%%%%%%%%%
%%%%%%%%%%%%%%%%%%%%%%%%%%%%%%%%%%%%%%%%

%%%%%%%%%%%%%%%%%%%%%%%%%%%%%%%%%%%%%%%%
%%%%%%%%%%%%%%%%%%%%%%%%%%%%%%%%%%%%%%%%

%%%%%%%%%%%%%%%%%%%%%%%%%%%%%%%%%%%%%%%%
%%%%%%%%%%%%%%%%%%%%%%%%%%%%%%%%%%%%%%%%

%%%%%%%%%%%%%%%%%%%%%%%%%%%%%%%%%%%%%%%%
%%%%%%%%%%%%%%%%%%%%%%%%%%%%%%%%%%%%%%%%

\section{Limitations and societal impacts}
\label{supa:impact}
This work aims to provide users with a method for efficient fine-tuning and controlled synthesis under diverse conditions. 
The tuning stage based on the pre-trained models while freezing the backbone network, so its transfer ability depends to a large extent on the performance of the upstream model. 
In addition, it generates results that meet expectations based on the training data and the specified conditional inputs supplied by the users. 
Conversely, the malicious utilization of high-risk data could potentially lead to the generation of misleading outcomes. 
This underscores the importance of ethical considerations in the deployment of generative models to prevent the propagation of harmfully biased or false information.

\end{document}